\documentclass{article}

\usepackage{microtype}
\usepackage{graphicx}
\usepackage{subfigure}
\usepackage{booktabs} % for professional tables

\usepackage{hyperref}

\usepackage[accepted]{icml2024}

\usepackage{amsmath}
\usepackage{amssymb}
\usepackage{mathtools}
\usepackage{amsthm}
\usepackage{multirow} 

\usepackage[capitalize,noabbrev]{cleveref}

\theoremstyle{plain}

\theoremstyle{definition}

\theoremstyle{remark}

\usepackage[textsize=tiny]{todonotes}

\icmltitlerunning{CasCast: Skillful High-resolution Precipitation Nowcasting via Cascaded Modelling}

\begin{document}

\twocolumn[
\icmltitle{CasCast: Skillful High-resolution Precipitation Nowcasting via \\ Cascaded Modelling  }



\icmlsetsymbol{equal}{*}

\begin{icmlauthorlist}
\icmlauthor{Junchao Gong}{sch,yyy}
\icmlauthor{Lei Bai}{yyy}
\icmlauthor{Peng Ye}{yyy}
\icmlauthor{Wanghan Xu}{yyy}
\icmlauthor{Na Liu}{cma}
\icmlauthor{Jianhua Dai}{sms}
\icmlauthor{Xiaokang Yang}{sch}
\icmlauthor{Wanli Ouyang}{yyy}
\end{icmlauthorlist}

\icmlaffiliation{yyy}{Shanghai AI Laboratory}
\icmlaffiliation{sch}{Shanghai Jiao Tong University}
\icmlaffiliation{cma}{National Meteorological Information Center}
\icmlaffiliation{sms}{Shanghai Meteorological Service}

\icmlcorrespondingauthor{Lei Bai}{baisanshi@gmail.com}

\icmlkeywords{Machine Learning, ICML}

\vskip 0.3in
]



\printAffiliationsAndNotice{}

\begin{abstract}
Precipitation nowcasting based on radar data plays a crucial role in extreme weather prediction and has broad implications for disaster management. 
Despite progresses have been made based on deep learning, two key challenges of precipitation nowcasting are not well-solved: (i) the modeling of complex precipitation system evolutions with different scales, and (ii) accurate forecasts for extreme precipitation. 
In this work, we propose CasCast, a cascaded framework composed of a deterministic and a probabilistic part to decouple the predictions for mesoscale precipitation distributions and small-scale patterns. 
Then, we explore training the cascaded framework at the high resolution and conducting the probabilistic modeling in a low dimensional latent space with a frame-wise-guided diffusion transformer for enhancing the optimization of extreme events while reducing computational costs. 
Extensive experiments on three benchmark radar precipitation datasets show that CasCast achieves competitive performance. 
Especially, CasCast significantly surpasses the baseline (up to \textbf{+91.8\%}) for regional extreme-precipitation nowcasting.
\end{abstract}

\section{Introduction}
\label{intro}
Precipitation nowcasting based on weather radar data plays a vital role in predicting local weather conditions over a period of up to two hours~\cite{climaguidelines}. These predictions are essential for various social sectors, including energy management and traffic scheduling. Moreover, precipitation forecasting serves as a critical tool for warning and mitigating disasters such as heavy rainfall and flooding. As a result, achieving skillful precipitation nowcasting is of utmost importance and has gained significant attention from researchers. 
% Precipitation nowcasting, depending on weather radar data, refers to the prediction of precipitation over a period from the present to 6 hours ahead with local weather details ~\cite{climaguidelines}. It is crucial for social sectors such as energy management and traffic scheduling. Another significance of precipitation forecasting is to alert disasters including heavy rainfall, flooding, and others. 

\begin{figure}[t]
\begin{center}
\centerline{\includegraphics[width=\columnwidth]{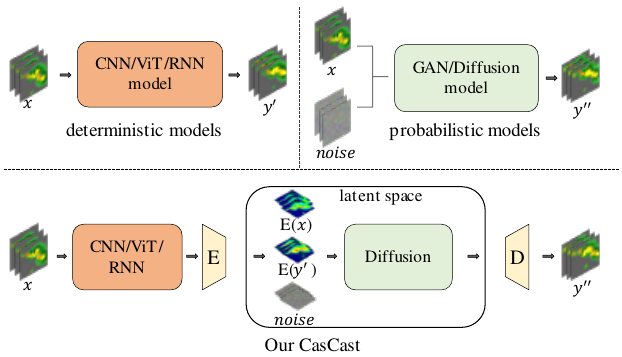}}
\vspace{-3mm}
\caption{Different precipitation nowcasting pipelines. Given context $x$, the prediction of deterministic models and probabilistic models are $y^{\prime}$ and $y^{\prime\prime}$, respectively. Our CasCast generates $y^{\prime\prime}$ conditional on $x$ and $y^{\prime}$ in the latent space. }
\vspace{-5mm}
\label{fig1}
\end{center}
% \vskip -0.1in
\end{figure}

Most precipitation events are caused by a combination of atmospherical systems with different scales ~\cite{prein2023multi}. On one hand, the mesoscale precipitation system evolves over spatial ranges of tens to hundreds of kilometers and time scales of several hours, driven and constrained by relatively stable large-scale circulation. On the other hand, the small-scale system, evolving within a range of a few kilometers and operating on time scales of minutes, is influenced by local processes such as heating, surface features, and other physical factors, which introduce stochasticity and unpredictability into its behavior. Thus, the combined influence of weather systems at multiple scales makes accurate prediction of precipitation events very challenging.

Deep learning methods have made significant contributions to precipitation nowcasting, with two broad categories emerging, i.e., deterministic models and probabilistic models. Deterministic models aim to capture the overall motion of the middle-scale precipitation system by providing a single-value prediction of the future state. However, these methods often encounter challenges related to blurriness and lack of fine-grained details.~\cite{shi2015convolutional, wang2017predrnn, guen2020disentangling, gao2022simvp, gao2022earthformer}. The reason is that deterministic models approximate the strong randomness, exhibited in the small-scale systems, by predicting mean values~\cite{ravuri2021skilful}. 
In contrast, probabilistic models use sampling from different latent variables to express the stochasticity of future weather systems, enabling them to capture small-scale weather phenomena~\cite{ravuri2021skilful, zhang2023skilful, gao2023prediff}. However, probabilistic methods may encounter challenges in accurately forecasting the large-scale distribution of precipitation that exhibits slow migration and high predictability, due to the stochastic modeling.
% for the middle-scale precipitation system that exhibits slow migration and high predictability. 
In summary, previous studies still struggle to predict the evolution of both the mesoscale and the small-scale systems simultaneously.
% To capture small-scale weather phenomena, probabilistic methods are utilized to express the stochasticity of future weather systems by sampling predictions from different latent variables ~\cite{ravuri2021skilful, zhang2023skilful, gao2023prediff}. However,  the probabilistic methods model the middle-scale precipitation system, which has properties of slow migration and high predictability, with stochastical modeling. This can lead to inaccuracies in forecasting the large-scale distribution of precipitation. 
% Previous studies struggle to simultaneously predict the evolution of both the mesoscale and the small-scale systems.

% In addition, accurate forecasting of extreme precipitation events is a critical requirement but often ignored in precipitation nowcasting
Another vital requirement of precipitation nowcasting is the forecasts for extreme precipitation events. 
Over the past 50 years, extreme-precipitation events have caused more than 1.01 million related deaths, and the economic losses are beyond US\$ 2.84 trillion~\cite{world2021wmo}. As 
% extreme weather processes typically exhibit lifetimes of tens of minutes and individual features at the convective scale (km-scale)
extreme weather processes typically exhibit short lifetimes of tens of minutes and involve convective-scale features at the kilometer scale~\cite{pulkkinen2019pysteps, ravuri2021skilful, zhang2023skilful}, effective forecasts for extreme precipitation require predicting the small-scale systems' evolutions at high resolution. Regrettably, existing methods have given less attention to this critical requirement.

% In addition to accurate predictions of general precipitation, forecasting extreme precipitation events is another crucial requirement in nowcasting. These events have had devastating consequences, causing over 1.01 million deaths and economic losses surpassing US\$2.84 trillion in the last 50 years~\cite{world2021wmo}. Extreme weather processes typically exhibit short lifetimes of tens of minutes and involve convective-scale features at the kilometer scale~\cite{pulkkinen2019pysteps, ravuri2021skilful, zhang2023skilful}. To effectively forecast extreme precipitation, it is essential to capture the evolutions of small-scale systems at high resolution. Regrettably, existing methods have given less attention to this critical requirement.

\begin{figure}[t]
\begin{center}
\centerline{\includegraphics[width=\columnwidth]{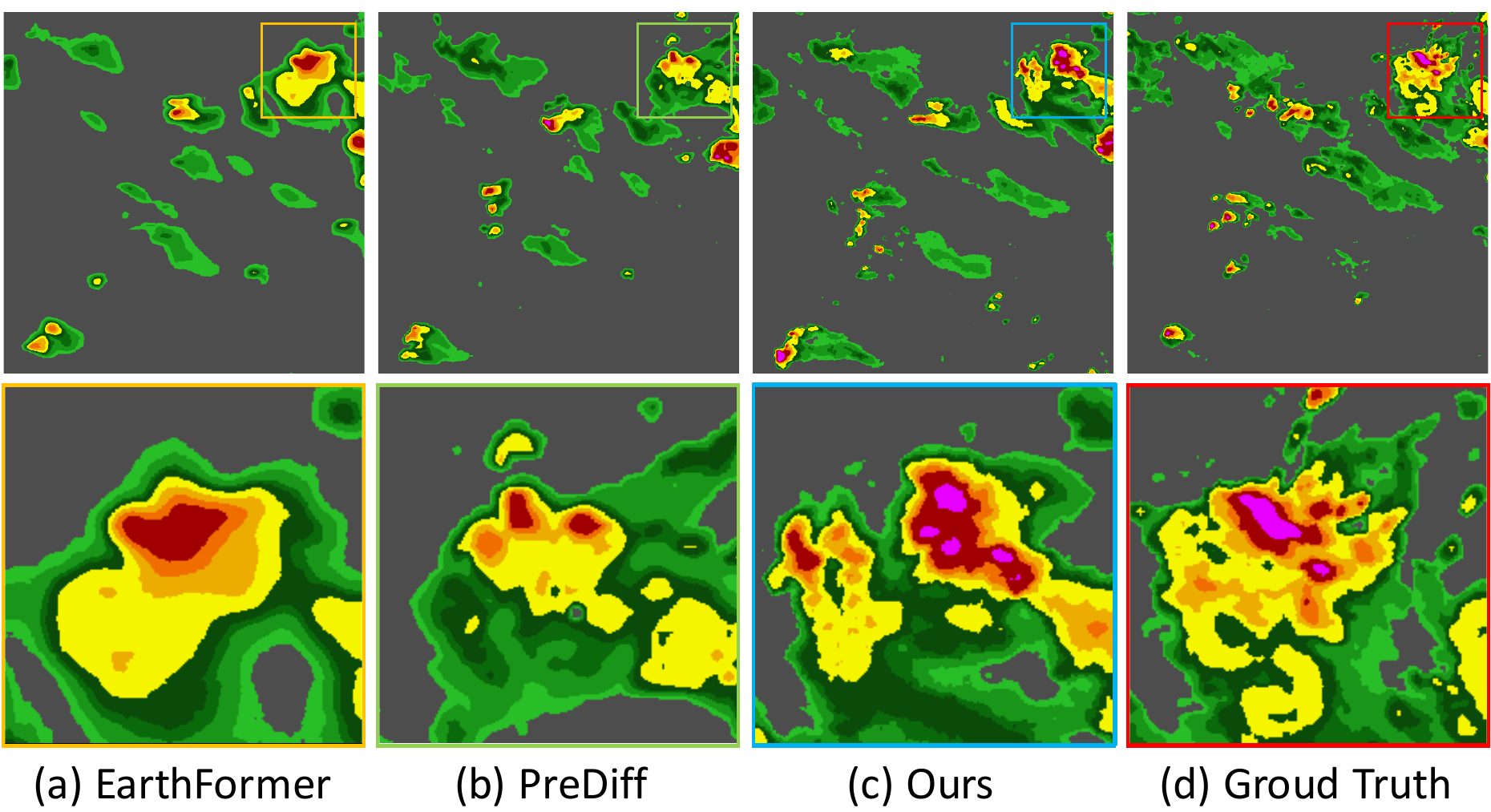}}
\vspace{-3mm}
\caption{Prediction visualization of different methods for a lead time of 60 minutes. The prediction of EarthFormer~\cite{gao2022earthformer} lacks small-scale patterns. The prediction of PreDiff~\cite{gao2023prediff} has a lower regional extreme value. In contrast, our CasCast method effectively captures both local patterns and regional extreme values.}
\vspace{-5mm}
\label{fig2}
\end{center}
% \vskip -0.1in
\end{figure}

In this work, we aim to achieve skillful high-resolution precipitation nowcasting with accurate global precipitation motion, realistic local patterns, and considerable regional extreme forecastings. We propose a novel cascaded nowcasting framework called CasCast, which disentangles precipitation nowcasting into predictions of mesoscale systems and generation of small-scale systems through a cascaded manner. As shown in ~\cref{fig1}, 
% the decomposition is achieved through a cascading manner. 
CasCast initially predicts the evolution of the mesoscale precipitation system with a deterministic model. Then, conditioned on the prediction of global precipitation distribution, a generative model is used for the generation of small-scale weather phenomena. Specifically, we choose diffusion models as the generative component of CasCast, as it does not suffer from mode collapse and artifacts commonly observed in GANs ~\cite{gao2023prediff}.
% Moreover, in order to enable the prediction of extreme precipitation events, we train the cascaded framework at high resolution, and further develop a frame-wise-guided diffusion transformer in a low-dimensional latent space to reduce computational costs and improve optimization.
Moreover, in order to enable the prediction of extreme precipitation events, we train the cascaded framework at high resolution, and further develop a frame-wise-guided diffusion transformer in a low-dimensional latent space. This frame-wise guidance in diffusion transformer ensures a frame-to-frame correspondence between blurry predictions and latent vectors, resulting in better optimization for the generation of small-scale patterns. The low-dimensional latent space reduces the computational cost caused by the high-dimensional characteristic of the radar data.
We test our CasCast on three classical radar precipitation datasets. On these datasets, our method achieves SoTA (state-of-the-art) and significantly improves the predictions for regional extreme precipitation nowcasting. As shown in ~\cref{fig2}, our method captures both the small-scale precipitation phenomena and forecasts a regional extreme event. 
% \bai{experiments, discuss figure 2, etc}
% for the generation of high-resolution local details which contributes to predicting extreme-precipitation events ~\cite{zhang2023skilful}.

Our contributions can be summarized as follows:
\vspace{-2mm}
\begin{itemize}
    % \item The first diffusion-based model that generates skillful precipitation nowcasting at a resolution of 1km. It has the capability to model small-scale phenomena and forecast regional extreme precipitation which is crutial for disaster warning.
    % \item We present CasCast, a cascaded model decoupling high-resolution precipitation nowcasting into the prediction of high-deterministic mesoscale systems and generation of high-stochastic small-scale systems.
    \item To predict the evolution of both mesoscale and the small-scale precipitation systems simultaneously, we introduce a novel cascaded  framework (CasCast), decoupling precipitation nowcasting into the prediction of high-deterministic part and the generation of component with strong stochasticity.
    % \item We introduce a diffusion transformer called CasFormer for the cascaded generation of high-resolution small-scale systems with frame-wise guidance in a latent space.
    \item We propose to train the cascaded framework at high resolution and generate forecasts with a frame-wise guided diffusion transformer (CasFormer) in a low-dimensional latent space, boosting the skill for extreme precipitation nowcasting.
    \item We conduct extensive experiments on three benchmark datasets, demonstraing the superior performance and robustness of the proposed method, especially for the extreme regional precipitation, e.g., outperforming the baseline model up to \textbf{+91.8\%} on the SEVIR dataset. 
    % \bai{on the xxx dataset when xxx}.
    %Our model is validated on three datasets, achieving State-of-The-Art(SoTA) performance in high-resolution precipitation nowcasting. Specifically, our model exhibits advancing accuracy for regional extreme precipitation.
\end{itemize}

\section{Related Work}
\subsection{Precipitation Nowcasting with Deep Learning}
Deep learning methods for precipitation nowcasting can be classified into deterministic and probabilistic. ConvLSTM ~\cite{shi2015convolutional} leverages convolution layers and LSTM (Long Short-Term Memory) cells to respectively extract spatial and temporal features of precipitation for deterministic forecasting. PredRNN enhances the capability of spatial-temporal modeling by splitting long-term and short-term memory cells ~\cite{wang2017predrnn}. PhyDNet decomposes the prediction into random motion and pde-guided motion, aiming to generate predictions that are consistent with physics ~\cite{guen2020disentangling}. Other deterministic models forecast short-term precipitation in a non-recurrence way by efficient space-time cuboid attention ~\cite{gao2022earthformer} or simple CNN ~\cite{gao2022simvp}. However, the predictions of deterministic models are blurred, as deterministic loss functions can drive the model to predict an average of multiple uncertain future small-scale precipitation. To predict small-scale weather phenomena, DGMR ~\cite{ravuri2021skilful} and NowcastNet ~\cite{zhang2023skilful} are trained adversarially, constraining the nowcasting to be close to the real precipitation distribution with spatial and temporal discriminators. However, GANs are known to suffer from training instability, making them prone to issues such as mode collapse and the generation of artifacts. Recent works ~\cite{gao2023prediff, zhao2024advancing} adopt diffusion to avoid shortages of GANs. However, due to the sensitivity of diffusion to model architectures and hyperparameters in high-dimensional spaces, as well as the computational cost of iterative sampling ~\cite{chen2023importance, pmlr-v202-hoogeboom23a}, recent diffusion-based precipitation methods are limited to low-resolution.

\subsection{Diffusion Models for High-Resolution Generation}
To achieve high-fidelity and detailed generation results, many studies have employed diffusion methods to generate high-resolution images or videos. LDM ~\cite{rombach2022high} leverages a pre-trained autoencoder to compress non-semantic information in images to obtain low-dimensional latent representations and then conducts training and sampling in a lower-dimensional latent space to avoid the expensive cost of iterative sampling and the instability in generating high-quality results in high-dimensional spaces. In addition to training diffusion models in the latent space for high-resolution generation, there is also research exploring the direct generation of high-resolution images in pixel space. IMAGEN ~\cite{saharia2022photorealistic} achieves excellent generation results by utilizing a multi-resolution cascaded generation approach. Other models achieve high-resolution generation through sophisticated network designs and noise strategies ~\cite{ho2022imagen, pmlr-v202-hoogeboom23a, gupta2023photorealistic}. However, these methods do not specifically focus on short-term forecasting. They either do not take into account the characteristics of the radar echoes or lack decoupled modeling of the multiscale physical processes involved in precipitation.

\begin{figure*}[t]
\begin{center}
\centerline{\includegraphics[width=2\columnwidth]{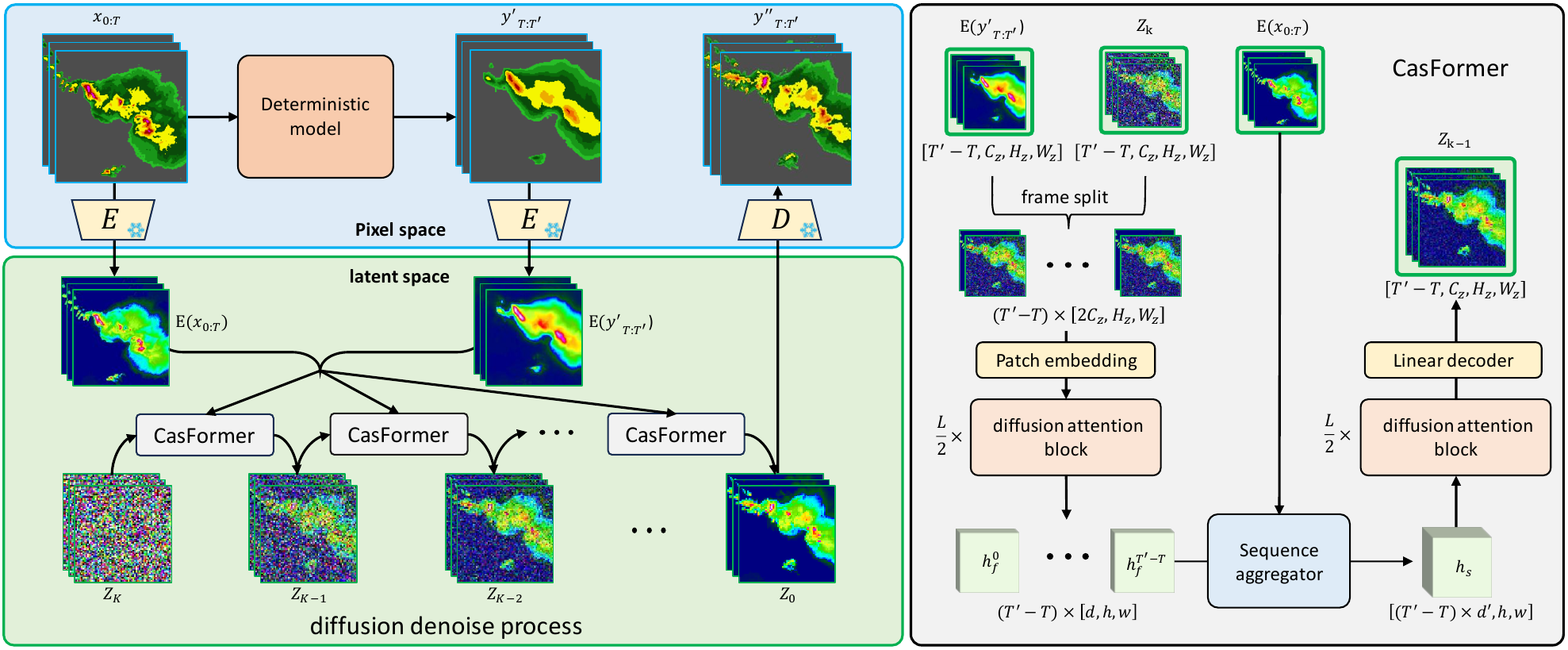}}
\vspace{-3mm}
\caption{\textbf{Left}: Overview of our CasCast. First, CasCast employs a deterministic model in pixel space to generate the blur prediction $y^{\prime}_{T:T^{\prime}}$ from previous observations $x_{0:T}$. Then, $x_{0:T}$ and $y^{\prime}_{T:T^{\prime}}$ are encoded into latent representations $\mathrm{E}(x_{0:T})$ and $\mathrm{E}(y^{\prime}_{T:T^{\prime}})$ by a pretrained frame-wise encoder $\mathrm{E}$. Last, conditioned on $\mathrm{E}(x_{0:T})$ and $\mathrm{E}(y^{\prime}_{T:T^{\prime}})$, the final prediction is generated through the diffusion denoise process on a novel CasFormer, and decoded back to the pixel space by a pretrained decoder. \textbf{Right}: Illustration of our CasFormer. First, $\mathrm{E}(y^{\prime}_{T:T^{\prime}})$ and the latent vector $z_{k}$ are split into framewise inputs and processed by patch embedding and $\frac{L}{2}$ layers of diffusion attention block. Then, frame-wise features $h^{0}_{f} \dots h^{T^{\prime}-T}_{f}$ and $\mathrm{E}(x_{0:T})$ are combined to the sequence-wise feature $h_s$ via a sequence aggregator, and used to predict the latent vector $z_{k-1}$. }
\vspace{-5mm}
\label{fig-CasCast}
\end{center}
% \vskip -0.1in
\end{figure*}

\section{Method}
\subsection{Task Formulation}
Precipitation nowcasting is commonly formalized as spatiotemporal forecasting problems of radar echoes ~\cite{shi2015convolutional, shi2017deep, veillette2020sevir}. Given the radar echo observations $x_{0:T}\in R^{T, C, H, W}$ from the past T time steps as the initial condition, short-time precipitation forecasting models' objective is to predict the radar echoes $y_{T:T^{\prime}} \in R^{T^{\prime}-T, C, H, W}$ of the future $T'-T$ frames. H and W represent the coverage range of radar data. The channel dimension C, with a value of 1, indicates the intensity of radar echoes.

\subsection{CasCaded Modelling}
As the evolution of most precipitation events is influenced by weather systems with different levels of randomness, we decouple the predictions of precipitation into the deterministic part and the probabilistic part in a cascaded manner as shown in ~\cref{fig-CasCast}. The deterministic model takes the previous observations $x_{0:T}$ as inputs to predict single-valued future precipitation $y^{\prime}_{T:T^{\prime}}$, formulated as $y^{\prime}_{T:T^{\prime}} =  \arg \max\limits_{y} p_{\theta_{d}}(y_{T:T^{\prime}}|x_{0:T})$, where $\theta_{d}$ is the parameter of deterministic part of CasCast. In CasCast, the architecture of the deterministic component can include most networks such as RNNs~\cite{shi2015convolutional}, CNNs~\cite{gao2022simvp} and transformers~\cite{gao2022earthformer}. We use MSE (mean square error) loss to train the deterministic model as 
\begin{align}
    L_{\theta_{d}}(y_{T:T^{\prime}}, y^{\prime}_{T:T^{\prime}}) = ||y_{T:T^{\prime}}-y^\prime_{T:T^{\prime}}||^2.
\end{align} The prediction $y^{\prime}_{T:T^{\prime}}$ suffers from blurriness of local details over time, because the deterministic loss drives models to predict the evolution of the samll-scale precipitation system into a mean value to represent the underlying uncertainty ~\cite{ravuri2021skilful}. However, $y^{\prime}_{T:T^{\prime}}$ effectively represents the global distribution of precipitation, because the mesoscale systems, dominating global patterns of the precipitation, are of high determinism. Then, The blurry prediction $y^{\prime}_{T:T^{\prime}}$  and initial condition $x_{ 0:T}$ are used as conditions for the generation of small-scale precipitation systems. The generative part samples latent Gaussian vector $z$ to capture the small-scale weather phenomena, described by
\begin{equation}
    z \sim N(0, 1), y^{\prime\prime}_{T:T^{\prime}}=\arg \max\limits_{y} p_{\theta_{p}}(y^{\prime}_{T:T^{\prime}}, x_{0:T}, z)
\end{equation}
where $\theta_{p}$ denotes the parameters of probabilistic models. Compared with pure probabilistic modelling ($y^{\prime\prime} = \arg \max\limits_{y} p_{\theta}(y|x_{0:T})$), the cascaded prediction avoids forecasting the evolution of highly deterministic parts in precipitation systems in a probabilistic manner, which contributes to the higher accuracy. Totally, CasCast disentangles the high-resolution precipitation nowcasting into two simpler tasks: predicting the global distribution of future precipitation and generating local weather patterns based on the blurry prediction.

\subsection{High-resolution Generation for Extreme Events}
% We employ diffusion in the latent space for high-resolution precipitation nowcasting generation in the probabilistic component of CasCast. Diffusion models do not exhibit mode collapse and artifacts compared with forecasting methods using GANs. Additionally, latent space enables efficient training and sampling in the diffusion-based precipitation nowcasting.

The probabilistic component of CasCast is used to improve the predictions for extreme precipitation events. Specifically, we apply a frame-wise guided diffusion transformer, which is conditioned by blurry predictions,  to generate high-resolution forecasts with local patterns in a low-dimensional latent space.

% \textbf{Preliminary of Diffusion Models.} Diffusion models learn the reverse process of gradually noising data $x_{0}$ into Gaussian noise. By progressively denoising the Gaussian noise, diffusion models generate samples $\hat{x} ~ \sim p(x_{0})$ where $p(x_{0})$ is data distribution. The noising process is a Markov process defined as $q(x_{k}|x_{k-1})=N(x_{k}; \sqrt{\alpha_{k}}x_{k-1}, \beta_{k}I), 1\leq k \leq K$ where $\beta_{k}=1-\alpha_{k}$ and $\alpha_{k}$ is  the $k^{th}$ step's constant coefficient of noise schedule. When $x_0 \sim p(x_{0})$ is given, the distribution of $x_{k}$ can be derived as $q(x_{k}|x_0)=N(x_{k}; \sqrt{\Bar{\alpha_k}}x_{0}, \sqrt{1-\Bar{\alpha_k}}I)$. In the denoising process, the joint distribution $p_{\theta}(x_{0:K})$ is factorized into multiplications of conditional distribution $p_{\theta}(x_{0:K}) = p(x_K) \prod_{i=K}^{i=1} p_{\theta}(x_{i-1}|x_{i})$. Diffusion models, parameterized as $\theta$, iteratively estimate $x_{k-1}$ from $p_{\theta}(x_{k-1}|x_{k})$ using MAE (maximum likelihood estimation). Finally, Gaussian noise $x_K$ is denoised into data sample $x_0$.

We take diffusion as the probabilistic part for its stable training process and remarkable generation ability. Diffusion models learn the reverse process of gradually noising data $x_{0}$ into Gaussian noise. By progressively denoising the Gaussian noise, diffusion models generate samples $\hat{x} ~ \sim p(x_{0})$ where $p(x_{0})$ is data distribution. The noising process is a Markov process defined as $q(x_{k}|x_{k-1})=N(x_{k}; \sqrt{\alpha_{k}}x_{k-1}, \beta_{k}I), 1\leq k \leq K$ where $\beta_{k}=1-\alpha_{k}$ and $\alpha_{k}$ is  the $k^{th}$ step's constant coefficient of noise schedule. When $x_0 \sim p(x_{0})$ is given, the distribution of $x_{k}$ can be derived as $q(x_{k}|x_0)=N(x_{k}; \sqrt{\Bar{\alpha_k}}x_{0}, \sqrt{1-\Bar{\alpha_k}}I)$. In the denoising process, the joint distribution $p_{\theta}(x_{0:K})$ is factorized into multiplications of conditional distribution $p_{\theta}(x_{0:K}) = p(x_K) \prod_{i=K}^{i=1} p_{\theta}(x_{i-1}|x_{i})$. Diffusion models, parameterized as $\theta$, iteratively estimate $x_{k-1}$ from $p_{\theta}(x_{k-1}|x_{k})$ using MAE (maximum likelihood estimation). Finally, Gaussian noise $x_K$ is denoised into data sample $x_0$.

% It is difficult to directly apply the high-dimensional radar data to the training and sampling of diffusion models. On the one hand, the radar echos, typically spanning several hours with a time interval of five minutes, form a long sequence. On the other hand, most radar data with a high resolution (such as 1km) cover areas within a radius of hundreds of kilometers ~\cite{shi2017deep, veillette2020sevir, Gwennaelle2020meteonet}. It is extremely challenging to directly generate them in such high-dimensional pixel space with diffusion. Another characteristic of radar data is redundancy. Radar echo usually contains redundant background intensity as the situation, where precipitation is continuous and consistent over a  coverage area of several hundred kilometers, is rare. The redundancy and high dimensionality exhibited in radar data motivate us to use a framewise autoencoder for the compression of radar data. We follow ~\cite{rombach2022high} to train a framewise autoencoder with pixel-wise loss and adversarial loss. Specifically, frame encoder $\mathrm{E}$ is trained to encode radar echo frame $x\in R^{1 \times C \times H \times W}$ into latent representation $\mathrm{E}(x) \in R^{1 \times C_z \times H_z \times W_z }$. The decoder reconstructs data frame $\hat{x}=\mathrm{D}(\mathrm{E}(x))$ from the latent representation. The latent space accelerates the training and sampling of diffusion models without severely corrupting the accuracy of prediction.

\textbf{Framewise Autoencoder.} It is difficult to directly apply diffusion models for precipitation nowcasting, because radar data, typically spanning several hours with a short time interval and covering areas within a radius of hundreds of kilometers at a high resolution (1km) ~\cite{shi2017deep, veillette2020sevir, Gwennaelle2020meteonet}, exist in a high-dimensional space. Another characteristic of radar data is redundancy. Radar echo usually contains redundant background intensity as the situation, when precipitation is continuous and consistent over an area of several hundred kilometers, is rare. The redundancy and high dimensionality exhibited in radar data motivate us to use a framewise autoencoder for the compression of radar data.  We follow ~\cite{rombach2022high} to train a framewise autoencoder with pixel-wise loss and adversarial loss. Specifically, frame encoder $\mathrm{E}$ is trained to encode radar echo frame $x\in R^{1 \times C \times H \times W}$ into latent representation $\mathrm{E}(x) \in R^{1 \times C_z \times H_z \times W_z }$. The decoder reconstructs data frame $\hat{x}=\mathrm{D}(\mathrm{E}(x))$ from the latent representation. The latent space accelerates the training and sampling of diffusion models without severely corrupting the accuracy of prediction.

\textbf{Framewise Guided Diffusion Transformer.}
We introduce CasFormer for the cascaded generation of precipitation nowcasting. As the generation is conditioned by the blurry deterministic prediction $y^{\prime}_{T:T^{\prime}}$, there is a frame-by-frame correspondence between  $y^{\prime}_{T:T^{\prime}}$ and the final prediction $y^{\prime\prime}_{T:T^{\prime}}$. Motivated by such correspondence, Our CasFormer generates local details with frame-wise guidance. Specifically, in the process of denoising latent vector $z_k$ to $z_{k-1}$, CasFormer performs frame splitting on the latent representation $\mathrm{E}(y^{\prime}_{T:T^{\prime}})$ of blur prediction and latent vector $z_{k}$. During frame splitting, $z_k$ and $\mathrm{E}(y^{\prime}_{T:T^{\prime}})$ are first split along dimension T into independent frame-wise $\mathrm{E}(y^{\prime}_{T:T^{\prime}})^{j} \in R^{C_z, H_z, W_z}$ and $z_k^j \in R^{C_z, H_z, W_z}$ where $j$ indicates the $j^{th}$ frame in the prediction sequence. Then $\mathrm{E}(y^{\prime}_{T:T^{\prime}})^{j}$ and $z_k^j$  are concatenated along dimension C, resulting frame-wise input $\in R ^{2C_z, H_z, W_z}$. The frame-wise input is utilized to extract the frame-wise feature $h_f^j \in R^{d, h, w}$ with a patch embedding layer and $\frac{L}{2}$ layers of diffusion attention block ~\cite{peebles2023scalable}. Frame-wise features $h_f^0 \dots h_f^{T\prime-T}$ of the same prediction sequence are aggregated into the sequence-wise feature $h_s \in R^{(T^{\prime}-T)\times d^{\prime}, h, w}$ by a sequence aggregator. In the sequence aggregator, frame-wise features are combined as $h_s = MLP(concat(h_f^0 \dots h_f^{T\prime-T}))$, and $h_s$ are injected with information from observations $\mathrm{E}(x_{0:T})$ in the latent space by a cross attention block. In the decoding stage, $h_s$ are decoded into the latent vector $z_{k-1}$ of the $(k-1)^{th}$ step. Overall, CasFormer has a frame-wise encoding stage and a sequence-wise decoding stage. The frame-wise encoding provides better-matched conditions for each frame-wise latent vector, reducing the complexity of the denoising conditioned by a sequence of blurry predictions.  Sequence-wise decoding utilizes the sequence features from the sequence-aggregator to ensure the spatiotemporal consistency of precipitation nowcasting.

% As shown in ~\cref{fig-CasCast}, observation $x_{0:T}$ and the blurry prediction $y^{\prime}_{T:T^{\prime}}$ are encoded into latent representations $\mathrm{E}(x_{0:T})$ and $\mathrm{E}(y^{\prime}_{T:T^{\prime}})$ as the condition:
% \begin{equation}
% \begin{align}
%     % z_{cond}^{x} &= \varepsilon(x_{0:T}) \\
%     % z_{cond}^{y^{\star}} &= \varepsilon(y^{\star}_{T:T^{\prime}}) \\
%     % p_{\theta}(z_{0:K}|z_{cond}^{x}, z_{cond}^{y^{\star}}) &= p(z_{K}) \prod_{K}^{k=1} p_{\theta}(z_{k-1}|z_{k}, z_{cond}^{x}, z_{cond}^{y^{\star}})
%     z_{cond} &= [\mathrm{E}(y^{\prime}_{T:T^{\prime}}),  \mathrm{E}(x_{0:T})]
% \end{align}
% \end{equation}
 The training objective of CasFormer is the noise prediction loss ~\cite{ho2020denoising}:
\begin{equation}
\begin{aligned}
    % z_k &= \sqrt{\Bar{\alpha_k}} z^{y} + \sqrt{1-\Bar{\alpha_k}} \epsilon \\
    % L_{\theta_{p}}(z^{y}, z^{cond})&=E_{\epsilon, k}[||\epsilon - \epsilon_{\theta}(z_{k}, k, z_{cond}) ||_{2}^{2}] \\
    L_{\theta_{p}}(z^{y}, z^{cond})&=\mathbb{E}_{\epsilon, k}[||\epsilon - \epsilon_{\theta_{p}}(z_{k}, k, z_{cond}) ||_{2}^{2}] \\
\end{aligned}
\end{equation}
where $\epsilon \sim N(0,1)$, $k \sim U(0,1)$, $z^y=\mathrm{E}(y_{T:T^{\prime}})$, $z_k = \sqrt{\Bar{\alpha_k}}z^y  + \sqrt{1-\Bar{\alpha_k}} \epsilon$ and $z_{cond} = [\mathrm{E}(y^{\prime}_{T:T^{\prime}}),  \mathrm{E}(x_{0:T})]$.

\section{Experiment}
\subsection{Experimental Setting}
\subsubsection{Datasets}

To validate the ability of CasCast to generate skillful 1km-resolution precipitation, we conducted tests on three radar echo datasets including SEVIR~\cite{veillette2020sevir}, HKO-7~\cite{shi2017deep} and MeteoNet~\cite{Gwennaelle2020meteonet}.

\begin{table}[h]
% \vskip 0.1in
% \vspace{-3mm}
\caption{The detailed settings of different datasets. Interval refers to the time duration between consecutive radar frames. The model predicts $L_{out}$ frames conditioned on $L_{in}$ frames.  }
% \vspace{-3mm}
% \vskip 0.1in
\begin{center}
\resizebox{1\columnwidth}{!}{
\begin{tabular}{l|cccccccc}
\toprule
dataset & $N_{train}$ &  $N_{val}$ & $N_{test}$ & $resolution$ & size & interval & $L_{in}$ & $L_{out}$ \\ 
\midrule
SEVIR   & 35,718 & 9,060 &  12,159   & 1km          & 384   & 5min        & 13   & 12    \\
HKO7    & 8,772 & 492 & 1,152   &  2km          & 480    & 6min         &  10   & 10     \\
MeteoNet & 6,978 & 2,234 &  994     &  $0.01^{\circ}$   &   400   &  5min     &  12   &  12    \\ 
\bottomrule
\end{tabular}
}
\end{center}
\end{table}

\textbf{SEVIR.} SEVIR ~\cite{veillette2020sevir} is an Earth observation dataset that contains weather radar observations. The NEXRAD radar mosaic of Vertically Integrated Liquid (VIL) in SEVIR  can be used for short-term precipitation forecasting. VIL in SEVIR records some storm events and random events that occurred in the United States between 2017 and 2019. The coverage range of VIL is 384km x 384km with a resolution of 1km and a time interval of 5min. We follow ~\cite{gao2022earthformer} to split SEVIR into 35718 training samples, 9060 validation samples, and 12159 test samples. We benchmark the nowcasting of precipitation by predicting the future VIL for up to 60 minutes (12 frames), based on a 65-minute context of VIL (13 frames).

\textbf{HKO-7.} In HKO-7, the radar CAPPI reflectivity images have a resolution of 480×480 pixels and are taken from an altitude of 2km and cover a 512km × 512km area centered in Hong Kong ~\cite{shi2017deep}. The radar data are collected from 2009 to 2015 with a time interval of 6 minutes. We benchmark HKO-7 by predicting the future radar echo up to 60 minutes (10 frames) given 60-minute observation (10 frames), resulting in 8772 training samples, 492 validation samples, and 1152 test samples. 

\textbf{MeteoNet.} MeteoNet ~\cite{Gwennaelle2020meteonet} contains rain radar data from the northwest and southeast regions of France collected by Meteo France from 2016 to 2018. The radar in MeteoNet has a spatial resolution of 0.01°, and the time interval between observations is 5 minutes. We choose the radar observations from the southeastern region of France in MeteoNet and select a portion of 400x400 pixels from the top left corner to remove areas with low radar quality.

\subsubsection{Evaluation.} Following ~\cite{gao2023prediff}, we utilize the SSIM, Critical Success Index (CSI), Heidke Skill Score (HSS), and the Continuous Ranked Probability Score (CRPS) to evaluate the quality of precipitation nowcasting. HSS and CSI are computed on a per-pixel basis, enabling the identification of position inaccuracy (more details are included in the appendix). 
We also follow ~\cite{ravuri2021skilful, zhang2023skilful} to report the CSI at 4$\times$4 and 16$\times$16 max pool scaling to reflect the model's ability for local precipitation pattern and regional extreme precipitation prediction. CRPS evaluates the uncertainty modeling ability of different models by measuring the discrepancy between the predicted distribution and the true distribution ~\cite{gao2023prediff}. When the prediction collapses to a single value, such as in a deterministic model, CRPS degrades to the Mean Absolute Error (MAE). A high CSI represents a good match between the model's predictions and the ground truth data on a pixel-wise or region-wise basis. A low CRPS indicates that the predicted distribution is close to the true distribution.  The calculation of all metrics for each model is performed on an ensemble of 10 members.

\begin{table*}[t]
\centering
\vskip 0.1in
\caption{Comparison of CasCast with various deterministic and probabilistic models on SEVIR dataset. CSI-M is the mean CSI score across thresholds $[16, 74, 133, 160, 181, 219]$. POOL1 is CSI scores at the grid resolution. POOL4 and POOL16 indicate 4-grid aggregations and 16-grid aggregations, respectively. $\dag$: EarthFormer is evaluated using the official checkpoint~\cite{gao2022earthformer} to assess the CSI scores for regional extreme precipitation. $\star$: The model is trained on a downsampled dataset with a size of 128 since it is found that training on the original dataset does not yield high-quality predictions.
% The model is trained in a downsampled dataset with a size of 128, because it cannot generate predictions of high quality after 100k training steps when trained in the original dataset. 
}
\label{sevir-table}
\vskip 0.1in
\begin{center}
% \begin{small}
% \begin{sc}
%1.65
\resizebox{1.65\columnwidth}{!}{
\begin{tabular}{l|c|c|c|ccc|ccc|ccc}
\toprule
\multirow{2}{*}{model} & \multirow{2}{*}{CRPS$\downarrow$} & \multirow{2}{*}{SSIM$\uparrow$} & \multirow{2}{*}{HSS$\uparrow$} & \multicolumn{3}{c|}{CSI-M$\uparrow$} & \multicolumn{3}{c|}{CSI-181$\uparrow$} & \multicolumn{3}{c}{CSI-219$\uparrow$} \\ \cline{5-13}
 &  & & & \multirow{1}{*}[-0.2em]{POOL1} & \multirow{1}{*}[-0.2em]{POOL4} & \multirow{1}{*}[-0.2em]{POOL16} & \multirow{1}{*}[-0.2em]{POOL1}& \multirow{1}{*}[-0.2em]{POOL4} & \multirow{1}{*}[-0.2em]{POOL16} & \multirow{1}{*}[-0.2em]{POOL1}& \multirow{1}{*}[-0.2em]{POOL4} & \multirow{1}{*}[-0.2em]{POOL16} \\
\midrule
$ConvLSTM$ ~\cite{shi2015convolutional}   & 0.0264 & 0.7749& 0.5232 & 0.4102 & 0.4163 & 0.4475 & 0.2453 & 0.2525 & 0.2977 & 0.1322 & 0.1380 & 0.1734 \\
$PredRNN$ ~\cite{wang2017predrnn} & 0.0271 & 0.7497 & 0.5192 & 0.4045 & 0.4161 & 0.4623 & 0.2416 & 0.2567 &  0.3214 & 0.1331  & 0.1447 & 0.1909 \\
$PhyDNet$ ~\cite{guen2020disentangling}    &0.0253 & 0.7649& 0.5311 & 0.4198& 0.4226 & 0.4410 & 0.2526  & 0.2532 &  0.2782 & 0.1362  & 0.1359 &  0.1526 \\
$SimVP$ ~\cite{gao2022simvp}    &0.0259 & 0.7772& 0.5280 & 0.4153 & 0.4226 & 0.4530 & 0.2532  & 0.2604 &  0.3000 & 0.1338 & 0.1394 & 0.1685 \\
$EarthFormer^{\dag}$ ~\cite{gao2022earthformer}     & 0.0251 & 0.7756& 0.5411 & 0.4310 & 0.4319 & 0.4351 & 0.2622  & 0.2542 &  0.2562 & 0.1448 & 0.1409 &  0.1481
\\ \hline
$NowcastNet$ ~\cite{zhang2023skilful} & 0.0283 & 0.5696& 0.5365 & 0.4152 & 0.4452 & 0.5024 & 0.2495  & 0.2935 &  0.3725 & 0.1422  & 0.1874 &  0.2700 \\
$LDM^{\star}$ ~\cite{rombach2022high}   & 0.0208 & 0.7495& 0.4386 & 0.3465 & 0.3442 & 0.3520 & 0.1470  & 0.1391 &  0.1432 & 0.0671 &0.0655  &  0.0717 \\
$PreDiff^{\star}$ ~\cite{gao2023prediff}    & 0.0202 & 0.7648& 0.4914 & 0.3875 & 0.3918 & 0.4157 &  0.2076  & 0.2069 &  0.2264 & 0.1032 & 0.1051 & 0.1213 \\ 
\hline
\textbf{CasCast(ours)}    & \textbf{0.0202} & \textbf{0.7797}& \textbf{0.5602} & \textbf{0.4401} & \textbf{0.4640} & \textbf{0.5225} & \textbf{0.2879} & \textbf{0.3179} &  \textbf{0.3900} & \textbf{0.1851} & \textbf{0.2127} &  \textbf{0.2841}\\
\bottomrule
\end{tabular}
}
\vspace{-3mm}
\end{center}
\end{table*}

\subsubsection{Training details.} We follow ~\cite{gao2022earthformer} to train the deterministic models. The training of autoencoder is the same as ~\cite{rombach2022high} except that we utilize the  AdamW optimizer with the highest learning rate of 1e-4 and a cosine learning schedule. As for settings of diffusion, we apply the linear noise schedule with 1000 diffusion steps and 20 denoising steps for inference with DDIM ~\cite{song2020denoising}. Classifier free guidance ~\cite{ho2022classifier} is adopted for training and inference. In the diffusion part, our CasFormer is optimized by AdamW optimizer with a learning rate of 5e-4 and a cosine learning rate scheduler. The training of diffusion takes 200k steps (18 hours) on 4 A100s with a global batchsize of 32.

\subsection{Compared to State of the Arts}

\begin{table*}[t]
\centering
\vskip 0.1in
\caption{Comparison of CasCast with various deterministic and probabilistic models on HKO-7 and MeteoNet datasets. }
\label{hko7_meteonet-table}
\vskip 0.1in
\begin{center}
% \begin{small}
% \begin{sc}
\resizebox{2\columnwidth}{!}{
\begin{tabular}{l|c|ccc|ccc|c|ccc|ccc}
\toprule
\multirow{3}{*}{model} & \multicolumn{7}{c|}{HKO-7} & \multicolumn{7}{c}{MeteoNet} \\
\cline{2-15}
& \multirow{2}{*}{CRPS$\downarrow$} & \multicolumn{3}{c|}{CSI-M$\uparrow$} & \multicolumn{3}{c|}{CSI-185$\uparrow$} & \multirow{2}{*}{CRPS$\downarrow$} & \multicolumn{3}{c|}{CSI-M$\uparrow$} & \multicolumn{3}{c|}{CSI-47$\uparrow$} \\ 
\cline{3-8} \cline{10-15}
 &  & \multirow{1}{*}[-0.2em]{POOL1} & \multirow{1}{*}[-0.2em]{POOL4} & \multirow{1}{*}[-0.2em]{POOL16} & \multirow{1}{*}[-0.2em]{POOL1}& \multirow{1}{*}[-0.2em]{POOL4} & \multirow{1}{*}[-0.2em]{POOL16} & & \multirow{1}{*}[-0.2em]{POOL1}& \multirow{1}{*}[-0.2em]{POOL4} & \multirow{1}{*}[-0.2em]{POOL16} & \multirow{1}{*}[-0.2em]{POOL1}& \multirow{1}{*}[-0.2em]{POOL4} & \multirow{1}{*}[-0.2em]{POOL16}\\
\midrule
$ConvLSTM$   & 0.0257 & 0.4000 & 0.4084 & 0.4280 & 0.1569 & 0.1843 & 0.2472 & 0.0218 & 0.3008 & 0.3050 &0.3465 & 0.0982 &0.1091 &0.1588\\
$PredRNN$  & 0.0252 & 0.3996 & 0.4146 & 0.4398 & 0.1633 & 0.1981 &  0.2634 & 0.0214  & 0.2914 & 0.3003 &0.3402& 0.0823& 0.0990 & 0.1462\\
$PhyDNet$    &0.0245 & 0.4213& 0.4121 & 0.3846 & 0.1807  & 0.1768 &  0.1913 & 0.0216  & 0.3120 &  0.3124 & 0.3356 &0.1106 & 0.1157 & 0.1482\\
$SimVP$    &0.0248 & 0.4236 & 0.4195 & 0.4134 & 0.1881  & 0.1953 &  0.2233 & 0.0218 & 0.3017 & 0.3143 & 0.3577 & 0.0997 & 0.1134 & 0.1599 \\
$EarthFormer$ & 0.0251 & 0.4096 & 0.4003 & 0.3950 & 0.1729  & 0.1731 &  0.1935 & 0.0224 & 0.2831 &  0.2855 & 0.3154 & 0.0787 & 0.0872 & 0.1208\\ \hline
$NowcastNet$ & 0.0296 & 0.4234 & 0.4518 & 0.4724  & 0.2025 &  0.2607 & 0.3601 & 0.0277  & 0.2955 &  0.3232 & 0.3734 & 0.1236 & 0.1521 & 0.2115  \\
$LDM^{\star}$     & 0.0260 & 0.3045 & 0.2738 & 0.2764 & 0.0517  & 0.0605 &  0.0928 & 0.0209 &0.2131  &  0.2191 & 0.2369 & 0.0359 & 0.0407 & 0.0552 \\
$PreDiff^{\star}$     & 0.0244 & 0.3221 & 0.3152 & 0.3046 &  0.0788  & 0.0852 &  0.1113 & 0.0197 & 0.2546 & 0.2668 & 0.2935 & 0.0490 & 0.0594 & 0.0867\\ 
\hline
\textbf{CasCast(ours)}    & \textbf{0.0205} & \textbf{0.4267} & \textbf{0.4608} & \textbf{0.4938} & \textbf{0.2158} & \textbf{0.2772} &  \textbf{0.3653} & \textbf{0.0180} & \textbf{0.3156} &  \textbf{0.3650} & \textbf{0.4420} & \textbf{0.1204} & \textbf{0.1563} & \textbf{0.2357}\\
\bottomrule
\end{tabular}
}
\vspace{-3mm}
\end{center}
\end{table*}

To validate the quality of high-resolution precipitation forecasts generated by our CasCast, we selected five deterministic models and three probabilistic models for comparison. ConvLSTM ~\cite{shi2015convolutional}, PredRNN ~\cite{wang2017predrnn} and PhyDNet ~\cite{guen2020disentangling} are deterministic short-term precipitation forecasting models designed based on recurrent mechanisms. SimVP ~\cite{gao2022simvp} and EarthFormer ~\cite{gao2022earthformer} are non-recurrent deterministic models built upon CNN or spatial-temporal transformer architechures. PreDiff ~\cite{gao2023prediff} and LDM ~\cite{rombach2022high} are diffusion-based probabilistic models, while NowcastNet ~\cite{zhang2023skilful} is a GAN-based probabilistic model.

~\cref{sevir-table} presents the quantitative experimental results of CasCast on the SEVIR dataset. In this experiment, CasCast incorporates EarthFormer as the deterministic \textbf{baseline}. We summarize some advantages of CasCast based on the results. (i) CasCast demonstrates the lowest CRPS. CasCast exhibits at least a $19.6\%$ reduction in CRPS compared to deterministic models. CasCast achieves a significantly lower CRPS compared to the GAN-based probabilistic model NowcastNet. This suggests that the prediction distribution of CasCast is more similar to the data distribution. (ii) The CSI-M-POOL4 and CSI-M-POOL16 of CasCast, which assesses the model's ability to capture local pattern distributions~\cite{gao2023prediff}, are considerably higher than other models. Compared to the best-performing deterministic model, CasCast shows an improvement of $7.4\%$ in CSI-M-POOL4 and an improvement of $20.1\%$ in CSI-M-POOL16. This improvement indicates that CasCast is more capable of capturing local precipitation patterns.  (iii) CasCast demonstrates excellent ability in forecasting regional extreme precipitation. CSI-181-POOL16 and CSI-219-POOL16 reflect the accuracy for regional extreme precipitation with vertically integrated liquid $12.14 kg/m^{2}$ (181) and $32.23 kg/m^{2}$ (219) respectively, within a 16-kilometer range. Compared to PredRNN, which has the highest extreme precipitation forecasting scores among deterministic models,  CasCast exhibits a $21.3\%$ improvement in regional extreme nowcasting with a threshold value of 181, and a $48.8\%$ improvement with a threshold value of 219. Compared to the prediff model, which is also diffusion-based, CasCast achieves significantly improved accuracy in forecasting regional extremes. This is because CasCast directly generates forecasts at a 1km resolution, avoiding the issue of extreme value attenuation caused by upsampling in Prediff. 
% Additionally, compared to the deterministic component of CasCast (EarthFormer), CSI-219-POOL16 substantially increased by \textbf{91.8\%.} 
(iv) CasCast has the highest CSI-M-POOL1 and HSS score, which indicates that its pixel-wise predictions are more accurate. This can be attributed to the hybrid deterministic and probabilistic prediction in CasCast. The deterministic component of CasCast ensures accurate predictions of the high-deterministic mesoscale motion during precipitation events, while incorporating the diffusion part enhances CasCast's modeling capability for high-stochasticity small-scale precipitation phenomena.

CasCast demonstrates generalization capabilities across different short-term forecasting datasets, as indicated in ~\cref{hko7_meteonet-table}. CasCast utilizes SimVP as the deterministic backbone for both HKO-7~\cite{shi2017deep} and MeteoNet~\cite{Gwennaelle2020meteonet} datasets, as SimVP exhibits the best overall performance on these two datasets. CasCast achieves skillful high-resolution short-term forecasts on both the HKO-7 and MeteoNet datasets, showcasing its forecasting advantages as observed on the SEVIR dataset as well.

\subsection{Analysis and Ablation Study}
\subsubsection{Distortion Caused by Autoencoder}
\begin{figure}[t]
\begin{center}
\centerline{\includegraphics[width=\columnwidth]{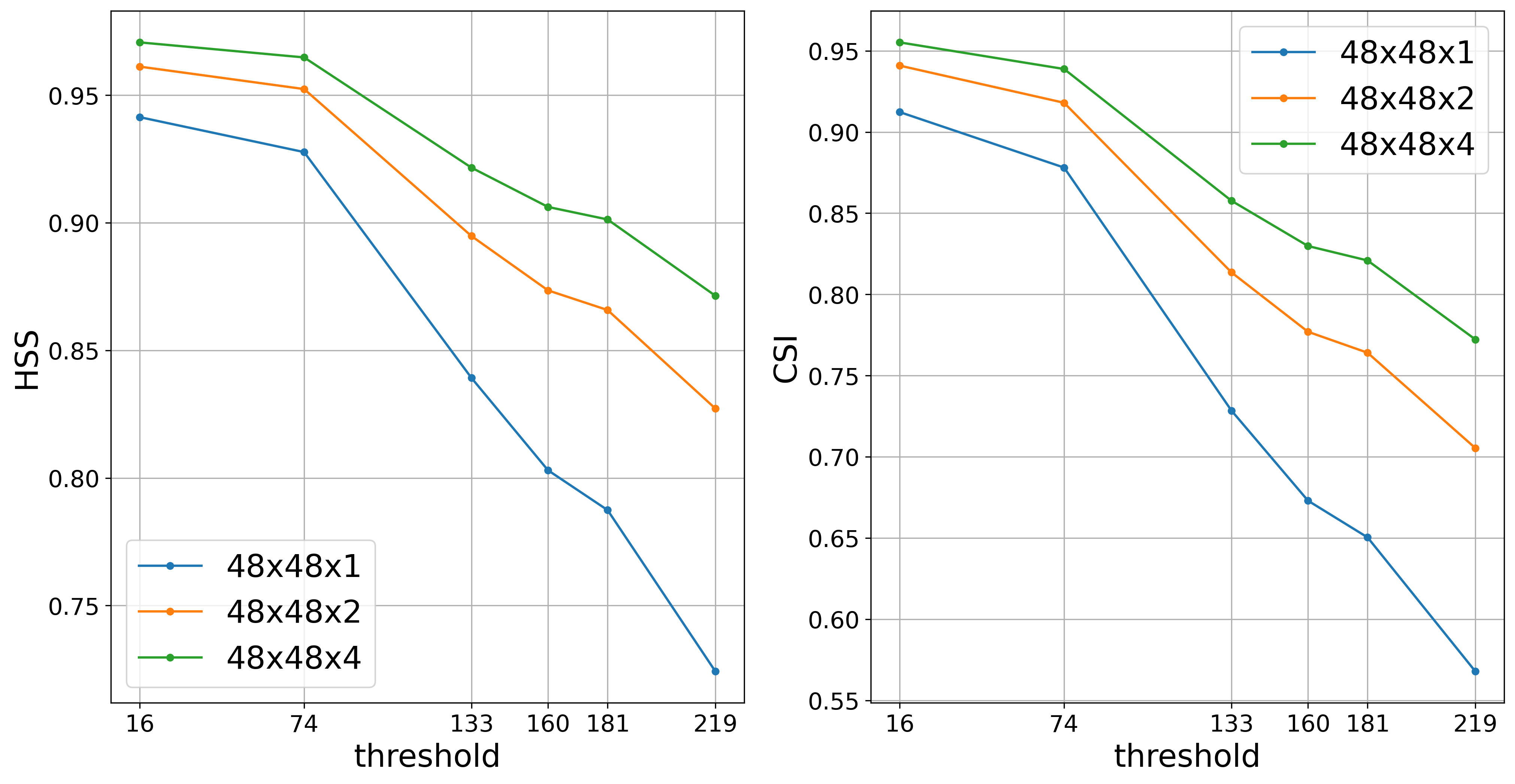}}
\vspace{-3mm}
\caption{HSS scores and CSI scores of autoencoders with different latent dimensions. Note that, $16, 74, 133, 160, 181, 219$ are different thresholds applied for computing the scores.}
\label{autoencoder-figure}
\end{center}
\vspace{-5mm}
\end{figure}

Due to the use of latent space, there is information loss caused by low-dimensional latent representation and the conversion process between pixel space and latent space. To investigate the influence, we measured the CSI and HSS scores between the original radar data and the reconstructed radar data with different latent dimensions. The experiment was conducted on the SEVIR dataset using the UNet ~\cite{rombach2022high} as the autoencoder. As exhibited in ~\cref{autoencoder-figure}, the CSI and HSS both decrease as the threshold increases. This implies that reconstructing high-intensity radar echoes is more challenging compared to low-intensity ones. When the last dimension of the latent space is doubled in size, the CSI and HSS increase, indicating a higher level of pixel-wise matching along with the increasing of latent space's dimension. However, the gain of increasing the dimension gradually diminishes especially for low thresholds. The experimental results indicate that there is still room to improve reconstruction performance by increasing the dimensions, especially for the extreme part.

\begin{figure*}[t]
\begin{center}
\centerline{\includegraphics[width=1.5\columnwidth]{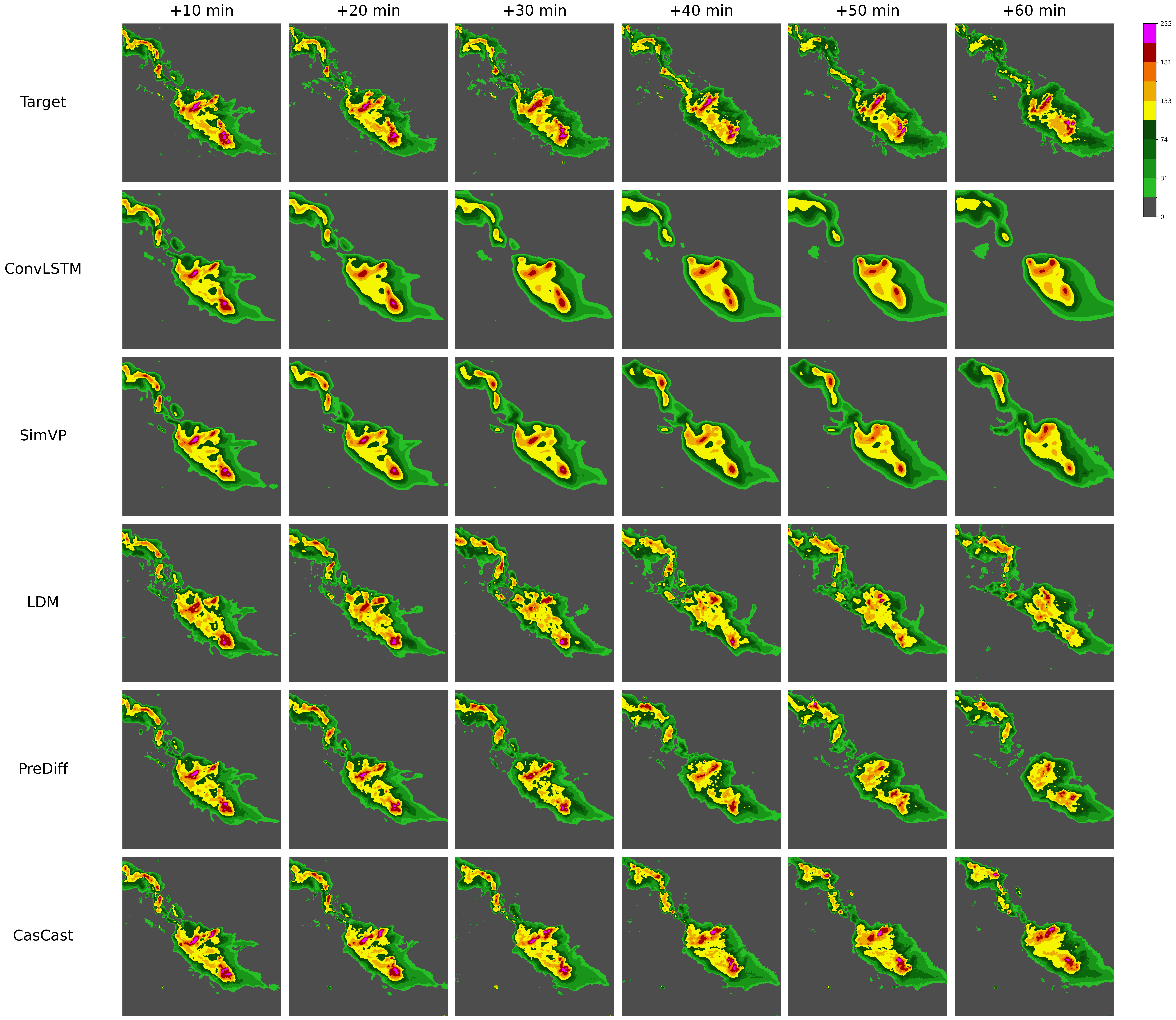}}
\caption{A set of example forecasts. From top to down denote Target, ConvLSTM, SimVP, LDM, PreDiff and CasCast (EarthFormer). From left to right denote forecasts of different lead times. (More qualitative results are shown in the Appendix.) }
\label{icml-historical}
\end{center}
\vspace{-5mm}
\end{figure*}

\subsubsection{Gains of Cascaded Modelling}
\begin{table}[t]
\vskip 0.1in
\caption{Seamless replacement of the deterministic part of CasCast on SEVIR dataset.}
\label{cascading-table}
\vspace{-3mm}
\vskip 0.1in
\begin{center}
% \begin{small}
\begin{sc}
\resizebox{\columnwidth}{!}{
\begin{tabular}{l|cc|cc}
\toprule
\multirow{2}{*}{model} &  \multicolumn{2}{c|}{CSI-M} & \multicolumn{2}{c}{CSI-219}  \\
\cline{2-5}
& \multirow{1}{*}[-0.2em]{POOL1} & \multirow{1}{*}[-0.2em]{POOL16} & \multirow{1}{*}[-0.2em]{POOL1} & \multirow{1}{*}[-0.2em]{POOL16} \\
\midrule
CasFormer & 0.3210 & 0.4402 & 0.0656 & 0.1733 \\
\hline
ConvLSTM   & 0.4102 & 0.4475 & 0.1322 & 0.1734 \\
CasCast(ConvLSTM) & 0.4145 & 0.5044 &  0.1575 &0.2593(+\textbf{49.54\%}) \\
\hline
SimVP & 0.4153 & 0.4530 & 0.1338 & 0.1685 \\
CasCast(SimVP) & 0.4206 & 0.5133 & 0.1629 & 0.2657(+\textbf{57.69\%}) \\
\hline
EarthFormer & 0.4310 & 0.4351 & 0.1448 & 0.1481 \\
CasCast(EarthFormer) & 0.4401 & 0.5225 & 0.1851& 0.2841(+\textbf{91.83\%}) \\

\bottomrule
\end{tabular}
}
\end{sc}
% \end{small}
\end{center}
\vspace{-3mm}
\end{table}

\begin{figure}[!h]
\begin{center}
\centerline{\includegraphics[width=\columnwidth]{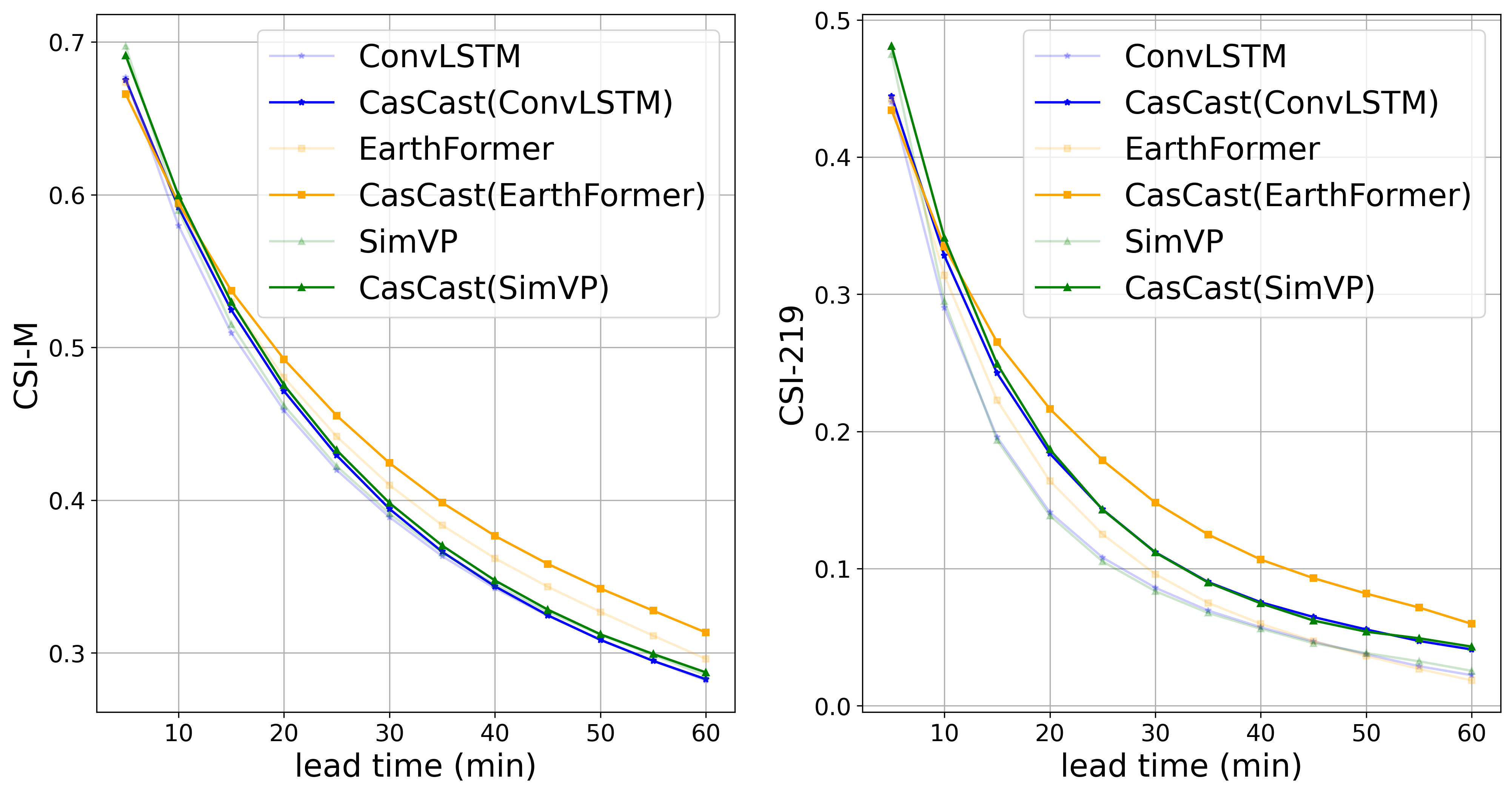}}
\vspace{-3mm}
\caption{Frame-wise CSI-M (left) and CSI-219 (right).}
\label{cascading-figure}
\end{center}
\vspace{-5mm}
\end{figure}

% To validate the effectiveness of the cascading strategy in CasCast, we conducted tests on the prediction performance of CasCast when using only the probabilistic generation model and when using different deterministic models within CasCast. 
% % To adapt to the generation without blur prediction as a condition, we made slight modifications to CasFormer. 
% When using only CasFormer for generation, the CSI-M is noticeably lower compared to deterministic models. This further confirms that when using either the probabilistic model or the deterministic model alone, there are either issues of pixel-wise mismatch or a limited ability to predict small-scale distribution and regional extreme values. Furthermore, the cascading strategy shows generalization across different deterministic components. When using ConvLSTM and EarthFormer as the deterministic parts of CasCast, the prediction results of CasCast are improved in both cases. Lastly, the performance of the deterministic model has an impact on the generation results of CasCast, especially on CSI-m. A better deterministic network will contribute to higher performance of CasCast.

To validate the effectiveness of the cascaded strategy in CasCast, we conducted tests on the SEVIR dataset when using only the probabilistic generation model and when using different deterministic models within CasCast (shown in ~\cref{cascading-table}, ~\cref{cascading-figure}).
The results further confirm that when using either the probabilistic model or the deterministic model alone, there are either issues of pixel-wise mismatch or a limited ability to predict small-scale distribution and regional extreme values.
When using ConvLSTM, SimVP, and EarthFormer as the deterministic parts of CasCast, the results have all improved. Especially, the CSI-219-POOL16 increases by \textbf{49.54\%}, \textbf{57.69\%}, \textbf{91.83\%} for each deterministic model, exhibiting considerable gain of cascaded modeling for regional extreme-precipitation nowcasting. 
Lastly, the performance of the deterministic model has an impact on the generation results of cascaded modeling. A better deterministic network will contribute to higher performance.

\subsubsection{Effectiveness of Frame-wise Guidance}
\begin{figure}[t]
\begin{center}
\centerline{\includegraphics[width=1\columnwidth]{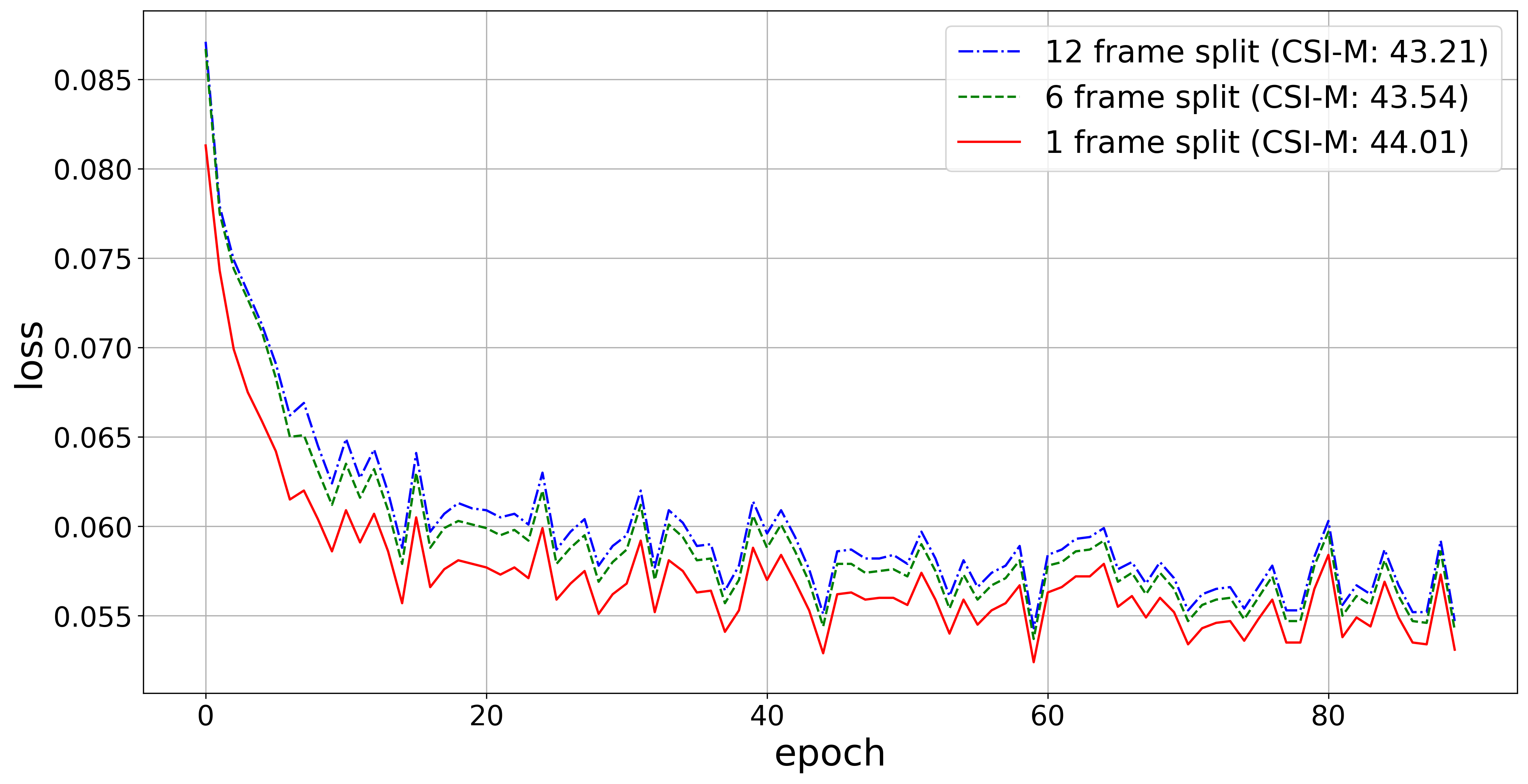}}
\vspace{-3mm}
\caption{The training loss of CasFormer with different frame splitting numbers in our CasCast.}
\label{framewise-figure}
\end{center}
\vspace{-5mm}
\end{figure}

We validate the effectiveness of frame-wise guidance in CasFormer on the SEVIR dataset. The results are exhibited in ~\cref{framewise-figure}. 1-frame-split is the default setting of CasFormer.  6-frame-split indicates dividing a sequence input into 2 inputs, each with 6 frames of blurry predictions as conditions.  When using a 12-frame-split setup, the encoding stage and decoding stage both take sequence-wise input, making CasFormer similar to the DiT model ~\cite{peebles2023scalable}. CasFormer with 1-frame-split outperforms its variants with 6-frame-split and sequence-wise-split in terms of optimization efficiency and final performance (CSI-M), which could be attributed to that 1-frame-split, ensuring a one-to-one correspondence between the latent vector $z$ and the blur prediction, provides the most matching conditions for frame generation.

% To validate the effectiveness of frame-wise encoding in CasFormer, we conducted tests on the SEVIR dataset to evaluate the performance of different frame-wise split in the encoder. 1frame-wise split is the setting of CasFormer. 6frame-wise split indicates a sequence input is split into 2 inputs with 6frames in encoding stage. When using a 12-frame-wise setup, where the 12-frame-wise encoder and sequence decoder both take sequence-wise inputs, CasFormer essentially becomes similar to the Diffusion transformer ~\cite{peebles2023scalable}. 1frame-wise splitting, which ensures a one-to-one correspondence between the latent vector $z$ and the blur prediction, provides the most matching conditions for frame generation. As a result, the 1frame-wise model shows the best convergence and achieves the highest CSI-m (0.8 higher than 12frame-wise model). 

\section{Conclusion}
In this paper, we propose a cascaded method (CasCast), decoupling the forecasts into predictions of global distribution and generation of local details,  for modeling complex precipitation systems with different scales.
To boost the accuracy of predictions for regional extreme events, we generate small-scale patterns at high resolution in a low-dimensional latent space with our CasFormer for efficient and effective cascading generation. 
 Experiments demonstrate that our methods predict realistic future radar echoes with high accuracy, especially for local extreme situations.

\textbf{Limitations and Future work}  CasCast achieves competitive performance on different datasets, but it requires retraining on these datasets which hinders its use in different regions and radar data. Although there are differences in terrain, climate, and data acquisition among radar echo datasets, the laws of precipitation exhibit commonalities. Our future work aims to explore how to predict precipitation with a unified model across different datasets.

\section*{Broader Impacts}
This research focuses on precipitation nowcasting which has an essential influence on urban planning and disaster management.
As an AI application for social good, our model boosts predictions for regional extremes. It improves disaster warnings such as storm warnings, flood forecasts, river level alerts, and landslide warnings, which helps the government, emergency agencies, and residents to prepare adequately and take necessary preventive and protective measures to minimize the losses and dangers caused by storms and floods.
However, there are also risks of inaccurate forecasts with mismatched locations or excessively high/low intensities. Such predictions may prevent the correct planning for social sectors such as energy management and traffic scheduling. Furthermore, generative forecasts are usually of multiple possibilities which require professionals to discern.

\nocite{langley00}

\bibliography{reference}
\bibliographystyle{icml2024}

%%%%%%%%%%%%%%%%%%%%%%%%%%%%%%%%%%%%%%%%%%%%%%%%%%%%%%%%%%%%%%%%%%%%%%%%%%%%%%%
%%%%%%%%%%%%%%%%%%%%%%%%%%%%%%%%%%%%%%%%%%%%%%%%%%%%%%%%%%%%%%%%%%%%%%%%%%%%%%%
% APPENDIX
%%%%%%%%%%%%%%%%%%%%%%%%%%%%%%%%%%%%%%%%%%%%%%%%%%%%%%%%%%%%%%%%%%%%%%%%%%%%%%%
%%%%%%%%%%%%%%%%%%%%%%%%%%%%%%%%%%%%%%%%%%%%%%%%%%%%%%%%%%%%%%%%%%%%%%%%%%%%%%%
\newpage
\appendix
\onecolumn
\section{More Details about Datasets.}
\subsection{SEVIR}
SEVIR dataset is a weather dataset containing observations from GOE satellites and  NEXRAD radar. We use the data of NEXRAD which are processed into radar mosaic of Vertically Integrated Liquid (VIL). Following ~\cite{gao2022earthformer}, the thresholds for the evaluation of the predictions for VIL data are set to be 16, 74, 133, 160, 181 and 219. To convert the thresholds $x$ into $R$ with the units of $kg/m^{2}$, which are the true units of VIL images, the following rule could be applied ~\cite{veillette2020sevir}:
% \begin{equation}
%     \begin{align}
%     R&=0 if x<=5 \\
%     R&=\frac{x-2}{90.66} if 5\textless x \leq 18 \\
%     R&=\exp{\frac{x-83.9}{38.9}} if 18 \textless x \leq 254
%     \end{align}
% \end{equation}

\[ 
R(x) =
\begin{cases}
    0, & \text{if } x\leq5 \\
    \frac{x-2}{90.66}, & \text{if } 5 \textless x \leq 18 \\
    \exp(\frac{x-83.9}{38.9}), & \text{if } 18 \textless x \leq 254 \\
\end{cases}
\]

\subsection{HKO7}
The raw radar echo images are generated by Doppler weather radar. The radar reflectivity values are converted  to rainfall intensity values(mm/h) using the Z-R relationship:
\begin{equation}
    dBZ=10\log a + 10b\log R
\end{equation} where R is ther rain-rate level, $a=58.53$ and $b=1.56$. The rainfall thresholds are $\geq0.5$, $\geq2$, $\geq5$, $\geq10$ and $\geq30$. According to Z-R relationship, dBZ thresholds are $30$, $52$, $66$, $77$, and $94$. Furthermore, the pixel thresholds $[84, 118, 141, 158, 185]$ are given by $pixel=\lfloor 255 \times \frac{dBZ+10}{70} + 0.5 \rfloor$ following ~\cite{shi2017deep}.

\subsection{MeteoNet}
MeteoNet is a dataset covering two geographical zones, North-West and South-East of France, during three years, 2016 to 2018. In our experiments, we use the rain radar part of the MeteoNet dataset. The region we use is the South-East of France. Original radar data is in the shape of $(565, 784)$, but some regions are missing in the data. We crop the data to keep the top-left portion with a size of $400 \times 400$. 

The thresholds used in MeteoNet are similar to HKO7 which evaluates the scores at Rain Rate (mm/h) $\geq0.5$, $\geq2$, $\geq5$, $\geq10$ and $\geq30$, respectively. As the radar records reflectivity factor $Z$ in decibels (dBZ), the rainfall rate $R$, expressed in millimeter per hour (mm/h), is approximated using the Marshall-Palmer relation:
\begin{equation}
    R = (\frac{Z}{200})^{\frac{1}{1.6}}
\end{equation}
in MeteoNet. It gives us thresholds in dBZ: 19, 28, 35, 40, 47.

\section{HSS and CSI}
HSS compares how often the pixel-wise predictions correctly match the ground truth. It can be calculated as :
\begin{equation}
    HSS = \frac{2(TP\times TN - FN \times FP) }{(TP+FN)(FN+TN)+(TP+FP)(FP+TN)}.
\end{equation}
The CSI (Critical Success Index) is another commonly used metric for measuring the accuracy of precipitation nowcasting. The definition of CSI is :
\begin{equation}
    CSI = \frac{TP}{TP + FN + FP}
\end{equation}

\section{More Visulization Results}

\begin{figure*}[t]
\begin{center}
\centerline{\includegraphics[width=0.8\columnwidth]{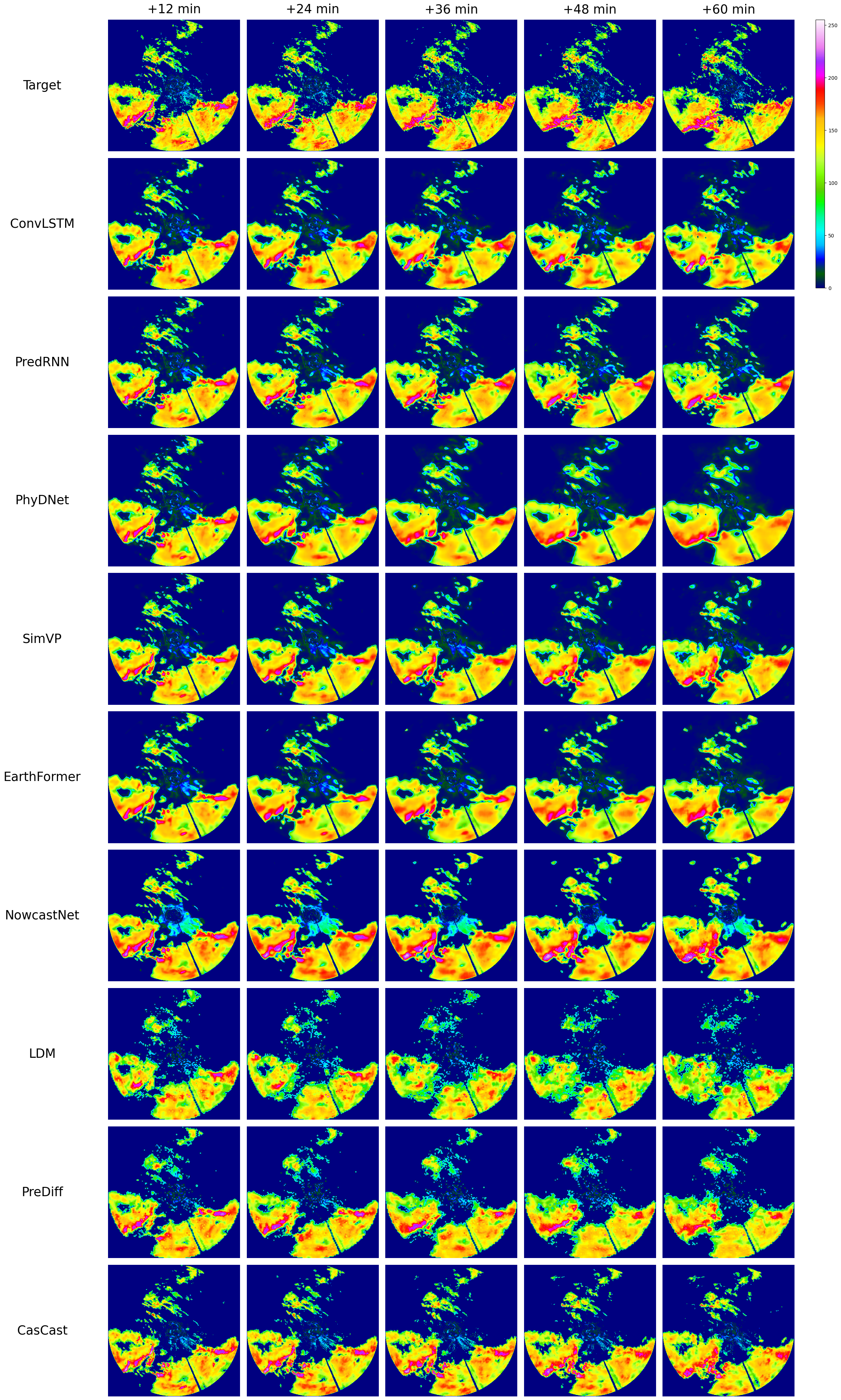}}
\caption{A set of example forecasts on HKO7. The deterministic component of CasCast is SimVP. }
\label{icml-historical}
\end{center}
\vspace{-3mm}
\end{figure*}

\begin{figure*}[t]
\begin{center}
\centerline{\includegraphics[width=0.8\columnwidth]{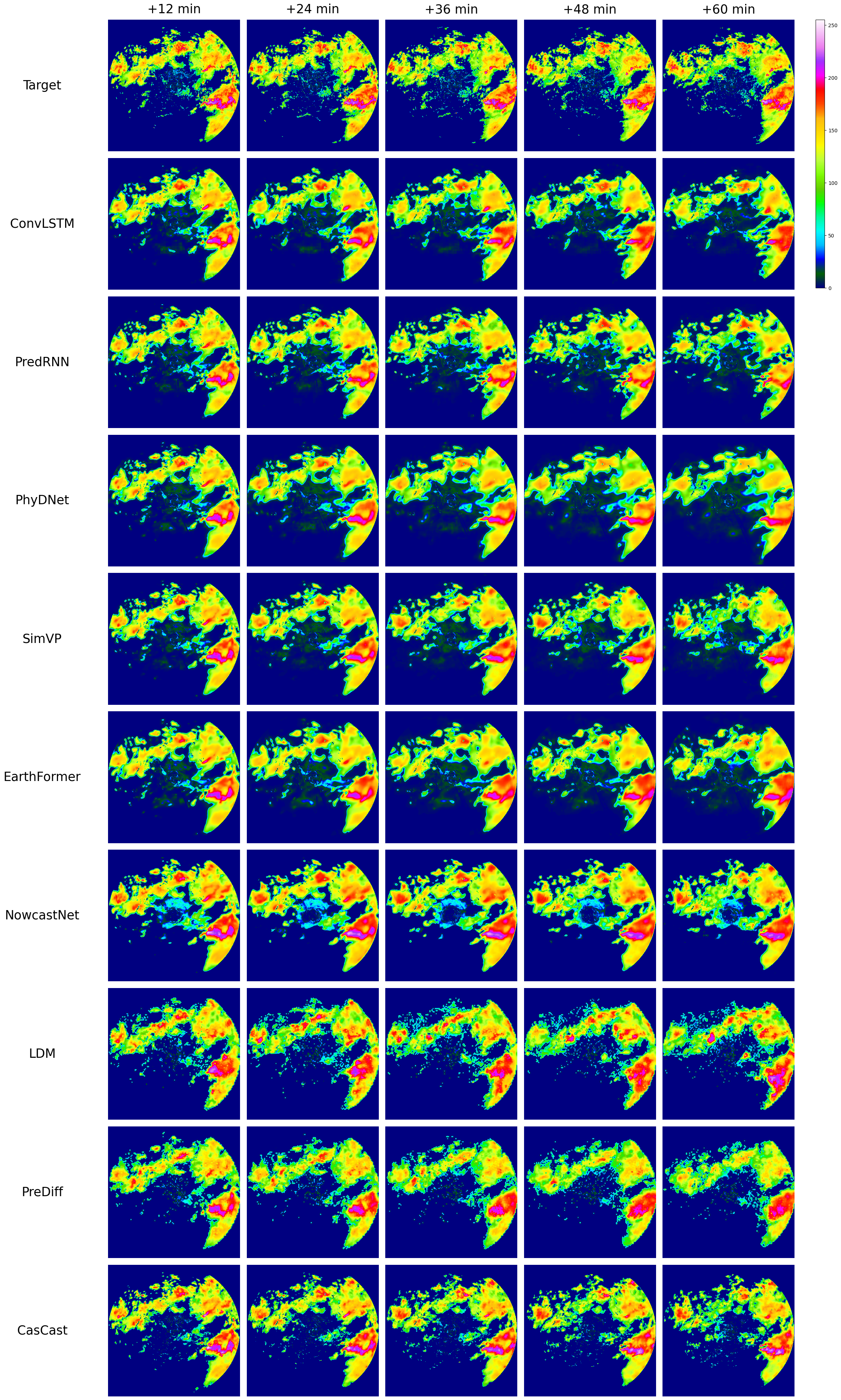}}
\caption{A set of example forecasts on HKO7. The deterministic component of CasCast is SimVP. }
\label{icml-historical}
\end{center}
\vspace{-3mm}
\end{figure*}

\begin{figure*}[t]
\begin{center}
\centerline{\includegraphics[width=0.8\columnwidth]{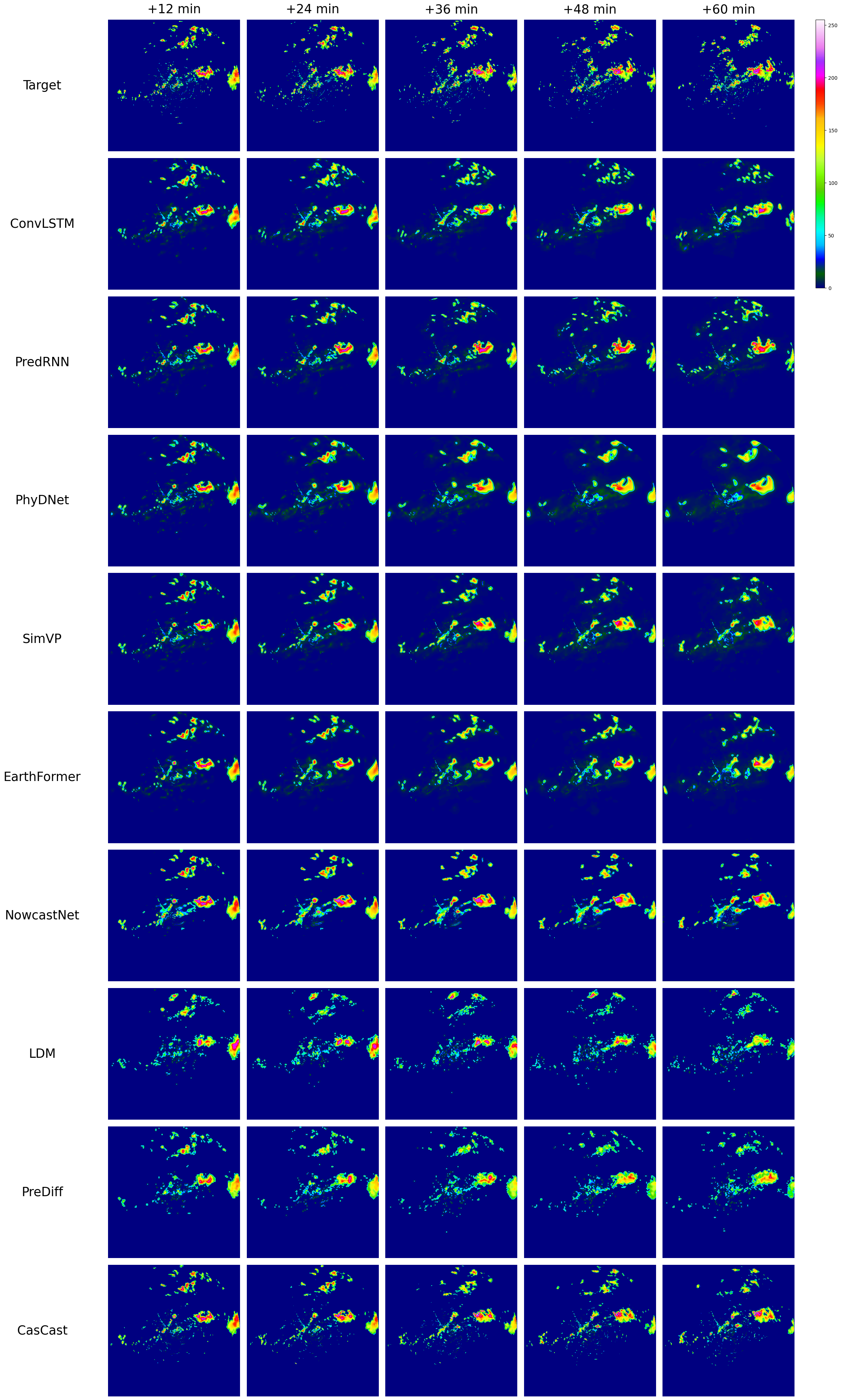}}
\caption{A set of example forecasts on HKO7. The deterministic component of CasCast is SimVP. }
\label{icml-historical}
\end{center}
\vspace{-3mm}
\end{figure*}

\begin{figure*}[t]
\begin{center}
\centerline{\includegraphics[width=0.8\columnwidth]{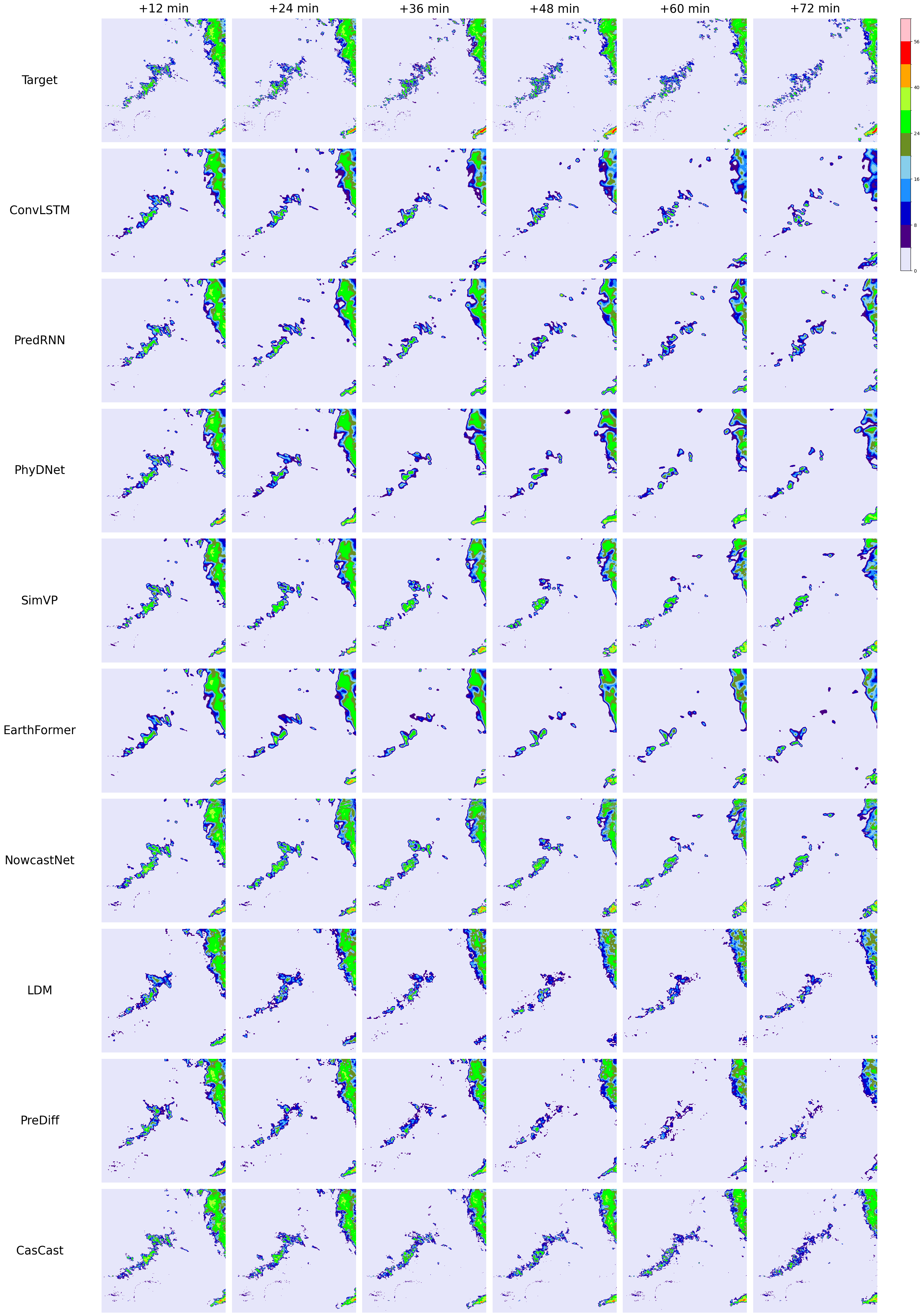}}
\caption{A set of example forecasts on MeteoNet. The deterministic component of CasCast is SimVP. }
\label{icml-historical}
\end{center}
\vspace{-3mm}
\end{figure*}

\begin{figure*}[t]
\begin{center}
\centerline{\includegraphics[width=0.8\columnwidth]{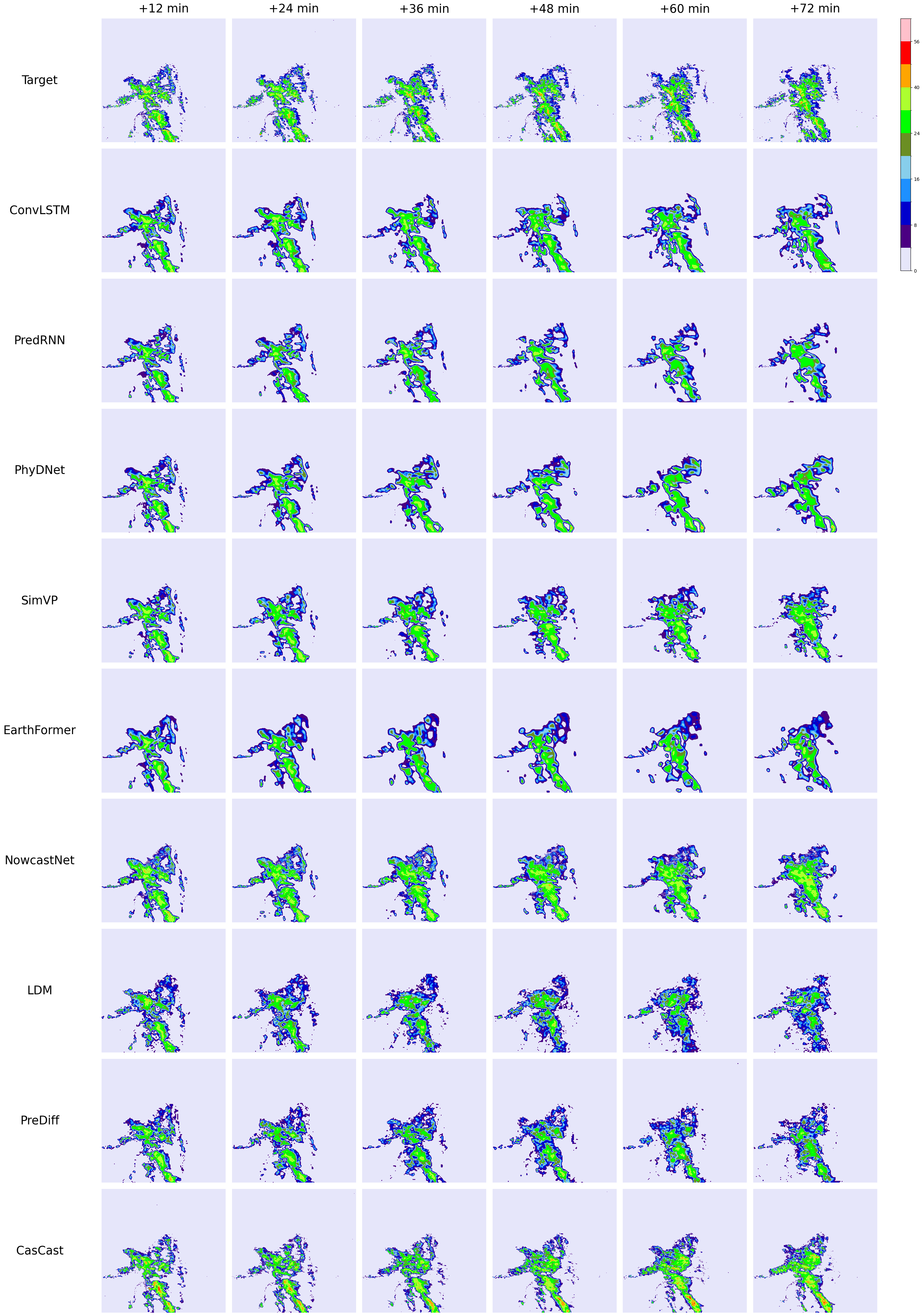}}
\caption{A set of example forecasts on MeteoNet. The deterministic component of CasCast is SimVP. }
\label{icml-historical}
\end{center}
\vspace{-3mm}
\end{figure*}

\begin{figure*}[t]
\begin{center}
\centerline{\includegraphics[width=0.8\columnwidth]{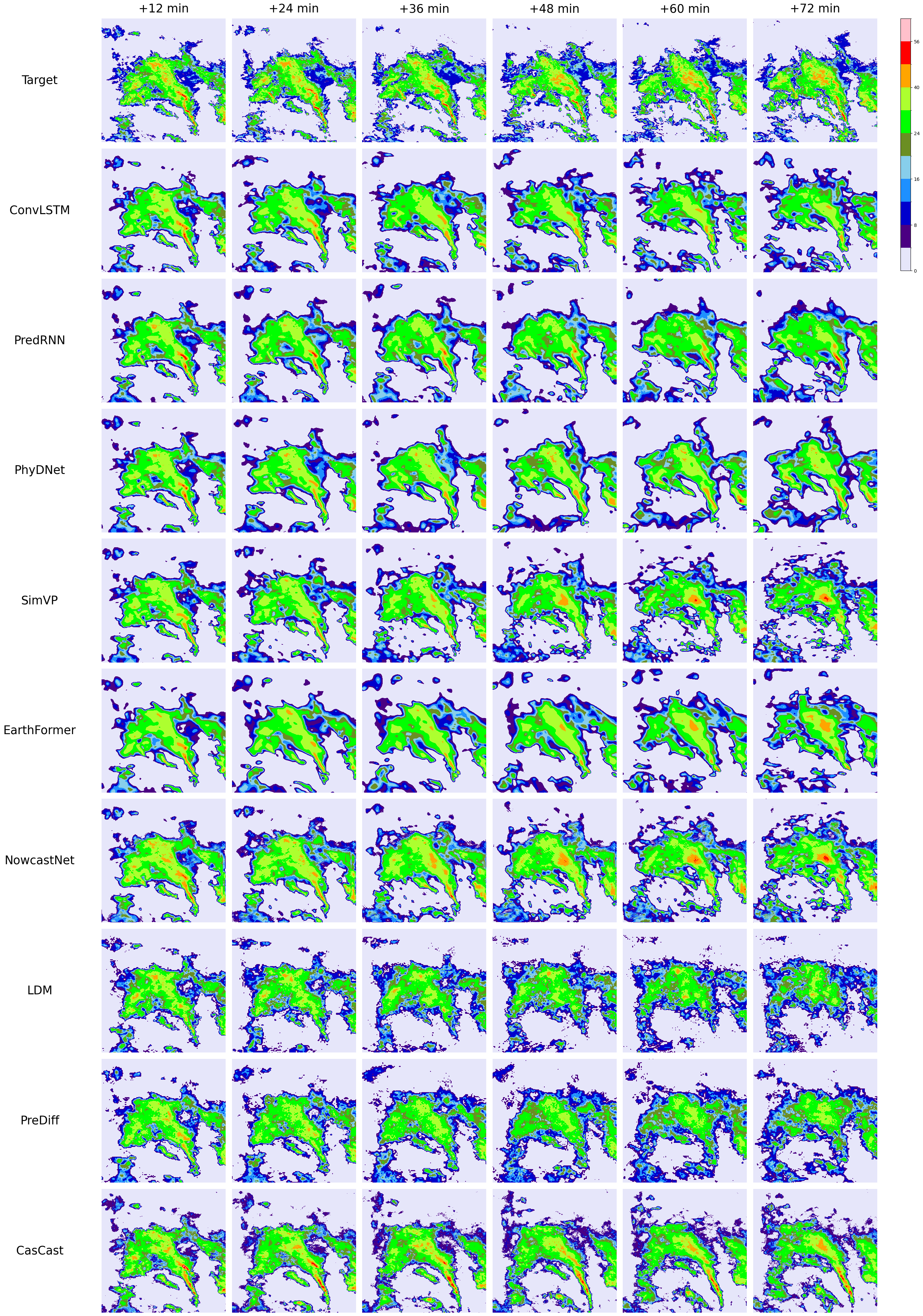}}
\caption{A set of example forecasts on MeteoNet. The deterministic component of CasCast is SimVP. }
\label{icml-historical}
\end{center}
\vspace{-3mm}
\end{figure*}

\begin{figure*}[t]
\begin{center}
\centerline{\includegraphics[width=0.8\columnwidth]{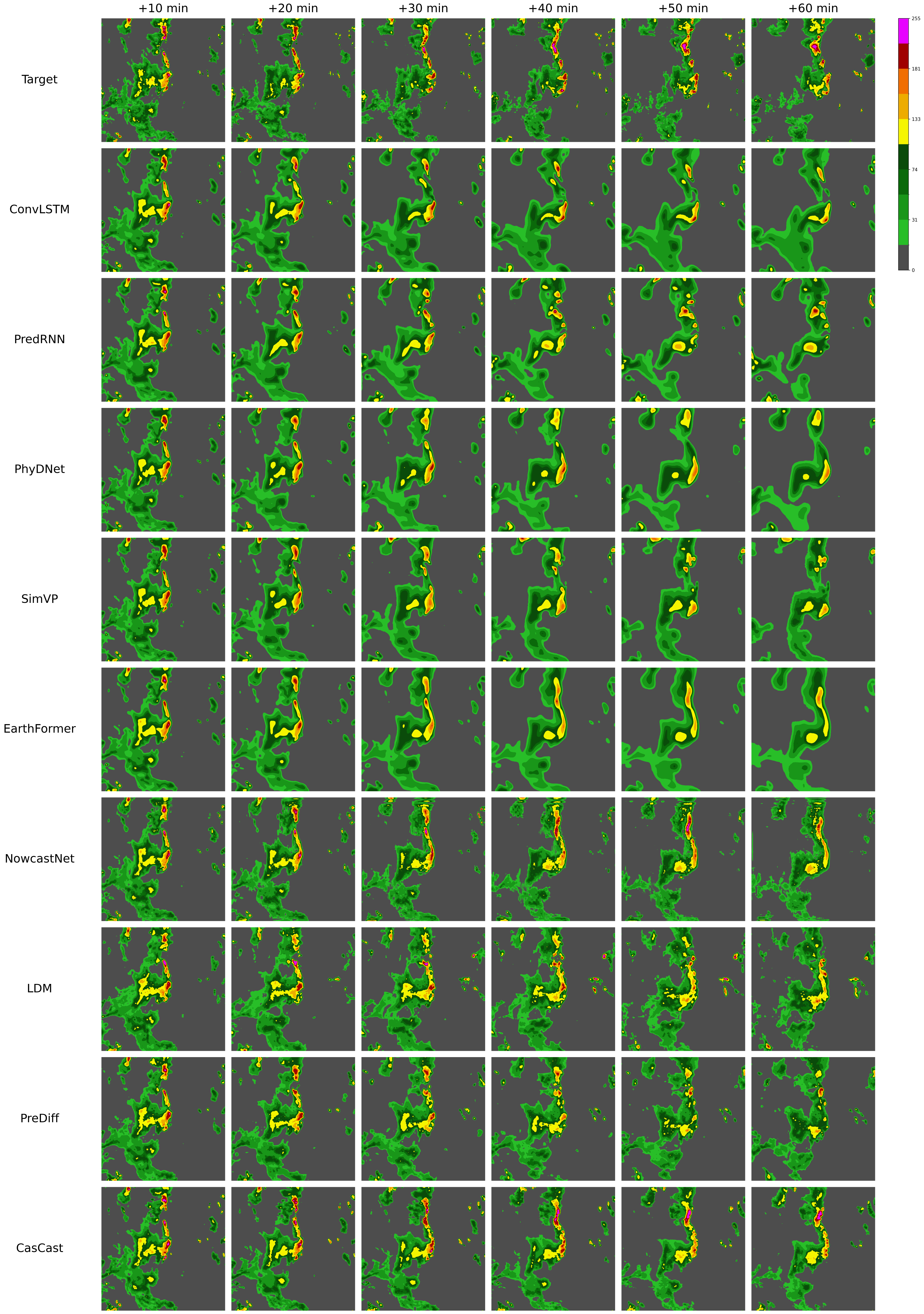}}
\caption{A set of example forecasts on SEVIR. The deterministic component of CasCast is EarthFormer. }
\label{icml-historical}
\end{center}
\vspace{-3mm}
\end{figure*}

\begin{figure*}[t]
\begin{center}
\centerline{\includegraphics[width=0.8\columnwidth]{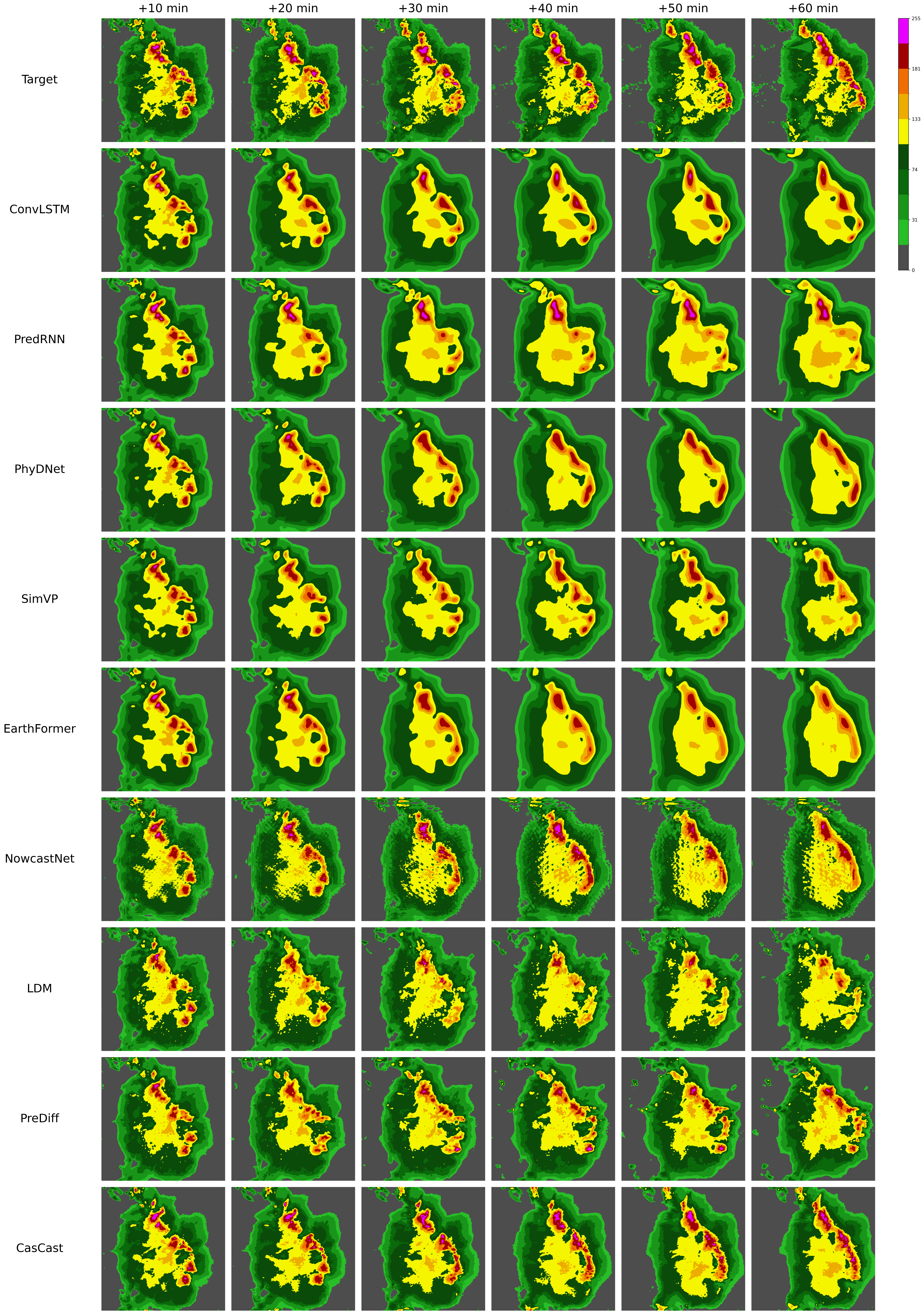}}
\caption{A set of example forecasts on SEVIR. The deterministic component of CasCast is EarthFormer. }
\label{icml-historical}
\end{center}
\vspace{-3mm}
\end{figure*}

\begin{figure*}[t]
\begin{center}
\centerline{\includegraphics[width=0.8\columnwidth]{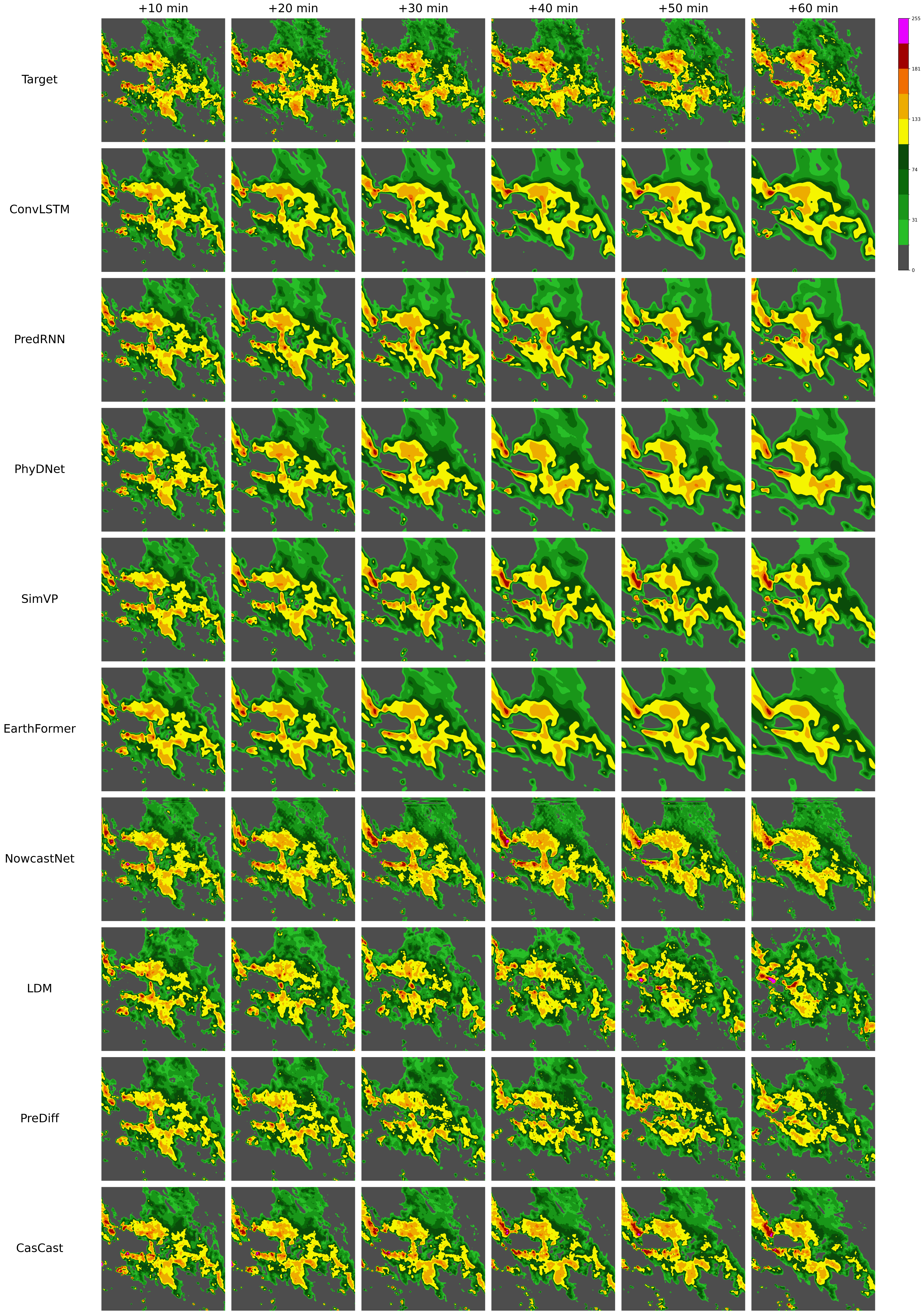}}
\caption{A set of example forecasts on SEVIR. The deterministic component of CasCast is EarthFormer. }
\label{icml-historical}
\end{center}
\vspace{-3mm}
\end{figure*}

%%%%%%%%%%%%%%%%%%%%%%%%%%%%%%%%%%%%%%%%%%%%%%%%%%%%%%%%%%%%%%%%%%%%%%%%%%%%%%%
%%%%%%%%%%%%%%%%%%%%%%%%%%%%%%%%%%%%%%%%%%%%%%%%%%%%%%%%%%%%%%%%%%%%%%%%%%%%%%%

\end{document}